\documentclass[journal]{IEEEtran}

\usepackage{amsthm}
\usepackage{moreverb,url}
\usepackage[colorlinks,bookmarksopen,bookmarksnumbered,citecolor=black,urlcolor=black]{hyperref}

\usepackage{color, mathrsfs}
\usepackage{enumerate}

\usepackage[shortlabels, inline]{enumitem}
\usepackage{graphicx}
\usepackage{amssymb}
\usepackage[normalem]{ulem}
\usepackage{bbm}
\usepackage{spverbatim} 
\usepackage{latexsym,amsbsy,amsfonts}
\usepackage[linesnumbered,ruled,vlined,algo2e]{algorithm2e}
\usepackage{caption}

\usepackage[caption=false]{subfig}
\usepackage[utf8]{inputenc}
\usepackage{amsopn,url,MnSymbol}

\usepackage{listings}
\newtheorem{thm}{Theorem}
\numberwithin{thm}{section}
\newtheorem{remark}[thm]{Remark}

\newtheorem{lemma}[thm]{Lemma}

\newtheorem{problem}{Problem}
\newcommand{\norm}[1]{\left\vert #1 \right \vert}	
\newcommand{\Norm}[1]{\left\Vert #1 \right \Vert}

\newtheorem{algorithm}{Algorithm}

\newtheorem{defi}[thm]{Definition}	
\newtheorem{assumptions}{Assumptions}

\newcommand{\Xfree}{X_{\text{free}}}
\newcommand{\Xsearch}{X_{\text{rand}}}

\newcommand{\R}{\mathbb R}

\newcommand{\st}{\ : \ }

\newcommand{\Cost}{\texttt{Cost}}
\newcommand{\Extend}{\texttt{Extend}}
\newcommand{\random}{\texttt{Random}}

\def\figtrajectory{./Figs/}

\begin{document}

\title{Probabilistic motion planning for non-Euclidean and multi-vehicle problems}

\author{Anton Lukyanenko
\thanks{A. Lukyanenko is with the Mathematics Department, George Mason University, Fairfax, VA 22030, alukyane@gmu.edu}, Damoon Soudbakhsh
\thanks{D. Soudbakhsh is with the Department
of Mechanical Engineering, Temple University, Philadelphia, PA 19122 USA, damoon.soudbakhsh@temple.edu
} 
}

\maketitle

\begin{abstract}
Trajectory planning tasks for non-holonomic or collaborative systems are naturally modeled by state spaces with non-Euclidean metrics. However, existing proofs of convergence for sample-based motion planners only consider the setting of Euclidean state spaces. We resolve this issue by formulating a flexible framework and set of assumptions for which the widely-used PRM*, RRT, and RRT* algorithms remain asymptotically optimal in the non-Euclidean setting. The framework is compatible with collaborative trajectory planning: given a fleet of robotic systems that individually satisfy our assumptions, we show that the corresponding collaborative system again satisfies the assumptions and therefore has guaranteed convergence for the trajectory-finding methods. Our joint state space construction builds in a coupling parameter $1\leq p\leq \infty$, which interpolates between a preference for minimizing total energy at one extreme and a preference for minimizing the travel time at the opposite extreme. We illustrate our theory with trajectory planning for simple coupled systems, fleets of Reeds-Shepp vehicles, and a highly non-Euclidean fractal space.
\end{abstract}

\begin{IEEEkeywords}
Nonholonomic Motion Planning, Motion, and trajectory Planning, Trajectory Planning for Multiple Mobile Robots, Cooperative Robots, and Multi-Robot Systems.
\end{IEEEkeywords}

\section{Introduction}
This paper presents a flexible axiomatic framework for probabilistic motion planning in a metric space setting that encompasses a variety of standard and novel scenarios.
The metric approach allows for generalizing the assumptions on the search problem. While previous convergence guarantees focused on single-vehicle holonomic robots, our approach allows us to provide convergence guarantees for PRM* and RRT* search tasks for individual or multiple holonomic or non-holonomic robots.

\subsection{Motion Planning}
A key question in robotic motion planning is determining the most efficient way a single robot or collaborative team of robots can reach their destinations in a complex environment. 
Even in the absence of obstacles, optimal trajectories are difficult to find. In the holonomic setting, it requires solving the geodesic equations with starting and ending constraints. In the non-holonomic setting, similar equations are available, but even the existence of well-behaved optimal trajectories  remains an area of active research see e.g.~\cite{MR1867362,jean_control_2014,Ardentov_2021, MR3958837}. Nonetheless, in many cases of interest, one is able to compute optimal obstacle-free trajectories (e.g.~Euclidean straight lines or Reeds-Shepp geodesics) in the state space.

In the presence of obstacles, exact trajectory planning is not computationally feasible, and  instead, sampling-based motion planning is employed. In this approach, one builds a discretization of the unobstructed space using a sequence of random or deterministic samples and then uses graph-theoretic methods to identify optimal trajectories within the discretization.
Optimizing this approach for single-vehicle trajectory planning has been the subject of many recent studies, including~\cite{varricchio_asymptotically_2018,janson_deterministic_2018,zhang_NovelLearning_2020,solovey_CriticalRadius_2020}.

The two primary types of probabilistic planning methods are based, respectively, on the Probabilistic Roadmaps (PRM) introduced by Kavraki et al.~\cite{kavraki_probabilistic_1996} and the Rapidly-exploring Random Trees (RRT) of Lavalle~\cite{lavalle_rapidly-exploring_1998}. Both types of algorithms sample a configuration space and build a graph-theoretic approximation of reachable destinations. PRM is optimized for multi-query applications, in which the starting and ending configurations may change between calls, and is performed in two stages: the roadmap phase that builds an approximating graph and a trajectory creation phase that identifies trajectories through the graph. On the other hand, RRT is optimized for single-query applications and builds a tree of possible trajectories rooted at the starting location and searching for points near the target location or region.
The core PRM and RRT algorithms have been shown to be effective at finding feasible trajectories in a wide variety of single-robot planning tasks~\cite{kavraki_probabilistic_1996, lavalle_rapidly-exploring_1998,lavalle_rapidly-exploring_2000, ranganathan_pdrrts_2004,lavalle_planning_2006,agha-mohammadi_sampling-based_2012,do_practical_2013, elbanhawi_sampling-based_2014}.

In \cite{karaman_sampling-based_2011}, Karaman and Frazzoli introduced two influential variants of the algorithms, called PRM* and RRT*, which provide asymptotically optimal trajectories and reduce the computational cost by limiting the distance at which samples are connected. 
Since then, probabilistic motion planning methods have been expanded to span many aspects of trajectory planning 
including the development of hybrid discrete-continuous algorithms~\cite{palmieri_rrt-based_2016}, 
addressing moving obstacles~\cite{choudhury_rrt*-ar_2013,otte_rrtx_2015}, 
improving nearest-neighbor searches for non-holonomic robotics~\cite{varricchio_efficient_2017,varricchio_asymptotically_2018}, improving search regions~\cite{denny2020dynamic,vonasek_SearchingMultiple_2020,tsao_sample_2020,xanthidis_MotionPlanning_2020,wang_EBRRTOptimal_2020}, and incorporating machine learning  methods~\cite{arslan2015machine,ekenna_adaptive_2016,zhang_NovelLearning_2020}.
The proof of convergence of  PRM* and  RRT*~\cite{karaman_sampling-based_2011} was extended in a sub-Riemannian setting to non-Euclidean systems in~\cite{karaman_sampling_2013}.

In 2019, an error was discovered in the analysis of RRT*, and an alternate proof of asymptotic optimality was provided for holonomic systems \cite{kleinbort_ProbabilisticCompleteness_2019,solovey2020revisiting}.
The same error appears in the analysis of RRT* in the sub-Riemannian setting \cite{karaman_sampling_2013} but is not addressed in the revised proof, requiring further analysis.

Our first contribution in this paper is presenting a flexible axiomatic framework for PRM* and RRT* in a metric space setting that encompasses both Euclidean and non-Euclidean systems, as well as more general search problems. In this setting, we provide convergence and asymptotic optimality guarantees for PRM*, RRT, and RRT*. The proof comes down to a direct analysis of RRT and an adaptation of the previous arguments for PRM* \cite{karaman_sampling_2013} and RRT*  \cite{solovey2020revisiting}.
\subsection{Multi-vehicle motion planning}

Motion planning for collaborative groups of robots, each of which is attempting to reach its destination while avoiding collisions, is a particularly challenging task, as the number of possible motions grows exponentially with the number of robots \cite{lavalle_planning_2006,shome_RoadmapsRobot_2021}. 
The computational time of RRT methods can be improved using biased or deterministic sampling~\cite{schmerling_optimal_2015,janson_fast_2015,denny_dynamic_2016,janson_deterministic_2018}. 
The \emph{sub-dimensional expansion} approach~\cite{solovey_FindingNeedle_2016, wagner2012probabilistic,wagner2015subdimensional} avoids this exponential growth by planning each robot's movements independently and then resolving any potential collision through local planning in a \emph{bubble space}.
Local trajectory planning is conducted after reducing the system's dimension at a local level to resolve conflicts. In such scenarios, the robots enter a box defined by the higher-order planner and leave it according to the poses defined by the planner.
In previous studies, local planning was based on robot task reassignment~\cite{wang2020shape}, or priority-based motion with holding patterns~\cite{turpin_capt_2014,tang2018complete,ma_searching_2019}, and localized RRT planning \cite{karlsson_multi-vehicle_2018}.

Our second contribution is to show that our  framework is compatible with the sub-dimensional expansion approach. That is, it can be used for multi-robot trajectory planning by combining the individual robots into a single state space while complying with the assumptions on the search space and maintaining convergence guarantees for PRM*, RRT, and RRT*. We also show that there are multiple choices of metric on the joint state space, which are controlled by a coupling parameter that influences the joint behavior of the fleet. When the fastest arrival to the destination is desired, the \emph{Manhattan metric} is appropriate. Alternately, minimization of total energy expenditure can be encoded using the \emph{supremum metric}. 
There are several ways to interpolate between these desires, as we demonstrate using the $\ell^p$ metrics (including the Euclidean metric when $p=2$).

\subsection{Outline of the paper}
In Section \ref{sec:theory}, we write down the data of a search problem, our geometric assumptions, and the PRM*, RRT, and RRT* algorithms. We then prove the probabilistic completeness of these algorithms and the asymptotic optimality of PRM* and RRT*. We finish by discussing multi-robot search problems, which are most naturally described using non-Euclidean metrics. We provide optimal geodesics in the obstacle-free state space for any choice of $p$, prove that the corresponding joint search problem satisfies our assumptions, and provide convergence guarantees for PRM*, RRT, and RRT*.

In Section \ref{sec:applications}, we demonstrate several search problems and show that they fit within our framework. We start with a simple case of holonomic robots, which allows us to demonstrate the effect of the coupling parameter on the search problem. We then confirm that the Reeds-Shepp vehicle model fits our assumption and demonstrate trajectory planning with corresponding  multi-vehicle systems. We finish by demonstrating the flexibility of our framework and convergence guarantee by performing RRT* trajectory planning in the non-manifold fractal setting of the Sierpinski gasket, where all basic notions in the framework and proof of  RRT* convergence have to be re-interpreted.

\section{Algorithms and Convergence Guarantees}
\label{sec:theory}
\subsection{Mathematical Preliminaries}
\label{sec:preliminaries}
A metric space is a set $X$ equipped with a \emph{metric function} $d: X\times X \rightarrow [0,\infty)$ that satisfies, for all $x,x',x''\in X$:
\begin{itemize}
    \item (non-degeneracy) $d(x,x')=0$ if and only if $x=x'$,
    \item (symmetry) $d(x,x')=d(x',x)$,
    \item (triangle inequality) $d(x,x'')\leq d(x,x')+d(x',x'')$.
\end{itemize}

In a metric space, one has (open) balls  $B(x,r)=\{x': d(x,x')<r\}$ and neighborhoods $N_r(A) = \{x: \max_{a\in A} d(a,x)<r\}$. 
A sequence $x_i$ in $X$ converges to a point $x_0$ if for any $\epsilon>0$ and sufficiently large $i$ one has that $d(x_i, x_0)<\epsilon$. A mapping $f: X\rightarrow Y$ between two metric spaces is \emph{continuous} if it preserves convergence: if a $x_i\in X$ converges to $x_0$ then the sequence $f(x_i)$ converges to $f(x_0)$.

A \emph{trajectory} (in mathematics literature, a \emph{path} or \emph{curve}) is a continuous function $\gamma: [a,b]\rightarrow X$, for some $a,b\in \R$.
The \emph{length} of $\gamma$ is given by 
$$\ell(\gamma) = \sup \sum_{i=0}^{n-1} d(\gamma(t_i), \gamma(t_{i+1}))$$
where the supremum is taken over all sequences $a=t_0<\ldots<t_i<\ldots<t_n=b$. The trajectory is a \emph{geodesic} if $d(\gamma(a),\gamma(b))=\ell(\gamma)$. Equivalently, if $\gamma$ is parametrized by unit speed, then $\gamma$ is a geodesic if and only if for any $a\leq s\leq t\leq b$ we have $d(\gamma(s)-\gamma(t))=s-t$. The space $X$ is \emph{geodesic} if any pair of points $x,x'\in X$ is connected by a geodesic.

A set $A\subset X$ is \emph{open} if for every point $a\in A$ there is a radius $r>0$ such that the open ball $B(a,r)$ is contained in $A$. A set $B$ is \emph{closed} if its complement is open. Both open and closed sets are examples of  \emph{Borel} sets, which are defined as elements of the smallest $\sigma$-algebra containing open sets, called the Borel $\sigma$-algebra, see, e.g., \cite{MR3098996} for details. A \emph{Borel measure} $\mu$ on $X$ is a function that assigns to every Borel set a size  $\mu(A)\geq 0$ (often interpreted as a volume or a probability), with the further restriction that $\mu(\emptyset)=0$ and that $\mu$ is countably additive on disjoint sets. 

Equipping a metric space with a Borel measure gives a \emph{metric measure space} $(X, d, \mu)$. In practice, metric measure spaces often satisfy additional conditions that relate the metric and the measure. A space is called \emph{$Q$-Ahlfors-regular on small scales} if there exist $r_0>0$ and $C>0$ such that for $r<r_0$  the measure of all balls $B(x,r)$ satisfies   $$\frac{1}{C} r^Q \leq \mu(B(x,r)) \leq C r^Q.$$

For example, the space $\R^n$ equipped with the Euclidean metric and the Lebesgue measure $\mathcal L^n$ (which formalizes the standard notion of volume) is $n$-Ahlfors-regular at all scales.

Given a random variable $\random$ defined on a subset $\Xsearch\subset X$ satisfying $0<\mu(\Xsearch)<\infty$, will say that $\random$ is \emph{uniformly distributed with respect to $\mu$} if for any Borel set $A\subset \Xsearch$ we have $$\mathbb P(\random \in A~|~\random \in \Xsearch) = \frac{\mu(A)}{\mu(\Xsearch)}.$$
Using product measures \cite{MR3098996}, one defines a corresponding probability for an event to occur given a (finite or infinite) sequence of identically distributed independent samples given by $\random$. An event is said to occur \emph{almost surely} if it occurs with probability 1.

\subsection{Assumptions, Algorithms, and Guarantees}
We phrase the probabilistic trajectory-finding problem as follows, using the terminology described in Section \ref{sec:preliminaries} and making additional assumptions about the search problem in Assumptions \ref{ass:basic}.

\begin{problem}
A search problem consists of:
\begin{enumerate}
    \item A metric-measure space $(X, d, \mu)$ called the \emph{configuration space},
    \item A \emph{search space} $\Xsearch\subset X$,
    \item An \emph{obstacle-free} space $\Xfree \subset \Xsearch$,
    \item A function \emph{\Extend} that connects pairs of points in $X$ by trajectories,
    \item A random variable \emph{\random} taking values in $\Xsearch$,
    \item Points $x_{start}, x_{end} \in \Xfree$,
    \item An error tolerance $\epsilon$, iteration constraint $n$, and connection radius sequence $\{r_i\}_{i=1}^\infty$.
\end{enumerate}
One then seeks a trajectory $\gamma: [a,b]\rightarrow \Xfree$ that is $\epsilon$-feasible or both $\epsilon$-feasible and $\epsilon$-optimal, meaning:
\begin{enumerate}
    \item ($\epsilon$-feasibility) $d(\gamma(a), x_{start})<\epsilon$ and $d(\gamma(b), x_{end})<\epsilon$,
    \item ($\epsilon$-optimality) if $\gamma'$ is any trajectory joining $x_{start}$ and $x_{end}$, then $\ell(\gamma)<\ell(\gamma')+\epsilon$.
\end{enumerate}
A problem is \emph{feasible} if a solution to the desired condition above in fact exists, and \emph{robustly feasible} if there exists a solution $\gamma$ that has, for some $\delta>0$,  strong $\delta$-clearance, i.e., $N_\delta(\gamma)\subset \Xfree$. (Note that in general, length-minimizing solutions do not exist, so we adjust the definition to be slightly different from the usual one, e.g. in \cite{karaman_sampling-based_2011}.)

A sampling-based algorithm is \emph{probabilistically complete} if for every $\epsilon$, the probability of generating an $\epsilon$-feasible trajectory limits to 1 as $n$ approaches $\infty$. It is furthermore \emph{asymptotically optimal} if for every $\epsilon$ the probability of generating an $\epsilon$-optimal trajectory limits to 1 as $n$ approaches $\infty$.
\end{problem}

We will extend PRM* and RRT* to non-Euclidean settings satisfying the following assumptions (see Section \ref{sec:preliminaries} for the definitions):
\begin{assumptions}\label{ass:basic}~
\begin{enumerate}
\item The search space $(X,d,\mu)$ is $Q$-Ahlfors-regular on small scales, for some $Q>0$,
\item For any $x,x'\in X$, the function $\Extend(x,x')$ provides a geodesic trajectory from $x$ to $x'$ parametrized at unit speed,
\item The search space satisfies  $0<\mu(\Xsearch)<\infty$,
\item $\random$ is uniformly distributed on $\Xsearch\subset X$ with respect to $\mu$.
\end{enumerate}
\end{assumptions}

We will use the following formulations of the RRT, PRM*, and RRT* algorithms.  

The PRM* algorithm builds a graph $(V, E)$ representing the search space. Optimal solutions within the graph can then be found using a graph-based algorithm such as A*.
\begin{algorithm}[PRM*]
Let $V$ consist of $n$ samples in $\Xfree$, selected using $\random$. For each pair of vertices $v_1, v_2\in V$, create an edge $(v_1, v_2)$ if $\ell(\Extend(v_1, v_2))\leq r_n$ and $\Extend(v_1,v_2)\subset \Xfree$.
\end{algorithm}

The RRT algorithm instead builds a tree  $(V, E)$ of solutions. Once a vertex $v$ near the destination is found, it can be traced back to the root to produce a corresponding trajectory. We will refer to the length of such a trajectory to the root as $\Cost(v)$.
\begin{algorithm}[RRT] Let $V=\{p\}$, $E=\emptyset$, and iterate until $V$ contains $n$ points, returning the resulting tree at the end. At each iteration:
    \begin{enumerate}
    \item Sample a point $x_{new}\in \Xfree$ using $\random$.
    \item Connect to a parent:
        \begin{enumerate}
        	\item Let $v_{nearest} = \operatorname{argmin}_{v\in V} d(v, x_{new})$.
        	\item If $d(v_{nearest}, x_{new})$ exceeds the maximal allowed travel distance, replace $x_{new}$ with an allowable point along $\Extend(v_{nearest},x_{new})$.
        	\item If $\Extend(v_{nearest},x_{new}) \subset \Xfree$, add the point $x_{new}$ to $V$ and add the edge $(v_{nearest}, x_{new})$ to $E$. Otherwise, proceed to the next iteration. 
        \end{enumerate}
    \end{enumerate}
\end{algorithm}

The RRT* algorithm interpolates between PRM* and RRT, producing an RRT-type tree  $(V,E)$ in the short-term while also incorporating a \emph{local} optimization process that allows it to build near-optimal trajectories from the root in the long-term, and in the process improve on the trajectories built by the RRT sub-algorithm.
\begin{algorithm}[RRT*] Let $V=\{p\}$, $E=\emptyset$, and iterate until $V$ contains $n$ points, returning the resulting tree at the end. At each iteration:
    \begin{enumerate}
        \item Sample a point $x_{new}\in \Xfree$ using $\random$.
        \item Make an initial connection:
            \begin{enumerate}
            	\item Let $v_{nearest} = \operatorname{argmin}_{v\in V} d(v, x_{new})$.
            	\item If $d(v_{nearest}, x_{new})$ exceeds the maximal allowed travel distance, replace $x_{new}$ with an allowable point along $\Extend(v_{nearest},x_{new})$.
            	\item If $\Extend(v_{nearest},x_{new}) \subset \Xfree$, add the point $x_{new}$ to $V$ and add the edge $(v_{nearest}, x_{new})$ to $E$. Otherwise, proceed to the next iteration. 
            \end{enumerate}
        \item Identify the locally-optimal parent: 
        \label{stepParent}
        \begin{enumerate}
            \item Let $A$ be the set of vertices $v \in V$ such that $d(v, x_{new})<r_{\norm V}$ and $\Extend(v, x_{new})\subset \Xfree$, excluding $x_{new}$.  
            \item If $A$ is empty, proceed to Step \ref{stepChildren}. 
            \item Let $x_{best} = \operatorname{argmin}_{a\in A} (\Cost(a)+\ell(\Extend(a,x_{new})))$. Remove the edge $(x_{nearest}, x_{new})$ from $E$ and add the edge $(x_{best}, x_{new})$.
        \end{enumerate}
        \item Identify the locally-optimal children:
        \label{stepChildren}
            \begin{enumerate}
            \item Let $A$ be the set of vertices $v \in V$ such that $d(x_{new}, v)<r_{\norm V}$ and $\Extend(x_{new},v)\subset \Xfree$, excluding $x_{new}$.  
            \item For each $a\in A$, if $\Cost(x_{new})+\ell(\Extend(x_{new}, a))<\Cost(a)$, 
            remove the edge from $a$ to its parent and add the edge $(x_{new}, a)$.
        \end{enumerate}
    \end{enumerate}
\end{algorithm}

\begin{thm}
\label{thm:generalized}
Consider a robustly feasible trajectory planning task satisfying Assumptions \ref{ass:basic}. Then, with probability 1:
\begin{enumerate}
    \item RRT provides a feasible solution,
    \item PRM* provides an asymptotically optimal solution if the connection radius satisfies  $r_i>\gamma \left(\frac{\log i}{i}\right)^{\frac{1}{Q}}$,
    \item RRT* provides an asymptotically optimal solution if the connection radius satisfies  $r_i>\gamma \left(\frac{\log i}{i}\right)^{\frac{1}{Q+1}},$ 
\end{enumerate}
where $\gamma>0$ is a parameter that does not depend on the choice of starting and ending points in the planning task.
\end{thm}
\begin{proof}[Sketch of Proof]
Let $\gamma$ be a solution to the search problem such that $N_\delta(\gamma)\subset \Xfree$. The idea is to approximate $\gamma$ by a sequence of samples along the trajectory, positioned in such a way that each algorithm has the opportunity to make the corresponding connections or pick better ones.

One can then use the methods of \cite{karaman_anytime_2011, solovey2020revisiting} to prove Theorem \ref{thm:generalized}, which were restricted to the holonomic (Euclidean) setting under our more general assumptions. Let us recall these methods for completeness.

For RRT, choose a sequence of balls $B(x_1, r), \ldots, B(x_M, r)$ of some small radius $r<\delta/4$ centered on $\gamma$ and covering all of $\gamma$. Assume also that $r$ is smaller than the maximum allowed travel distance. Each ball has a positive measure and, therefore, a positive probability of being sampled at each iteration.  Therefore, a point $x_1'$ will, almost surely, be eventually discovered in $B(x_1, r)$ and connected to $x_0$ or a closer point in the tree (note that the restriction $r<\delta/4$ together with the triangle inequality ensure that the connecting trajectory is collision-free). Subsequently, a point $x_2'$ will be sampled in $B(x_2, r)$ and connected to the tree (possibly connecting to $x_1'$). Continuing in this way, one eventually adds a point $x_M$ that is within distance $r$ of the destination.

Let us next sketch the proof for PRM*, following \cite{karaman_anytime_2011}.  For each time parameter $j$, one defines a radius $r'_j$, points $x_1, \ldots, x_{M_j}$ along $\gamma$, and a sequence of balls $B(x_1, r'_j), \ldots, B(x_{M_j}, r'_j)$ such that any pair of samples in adjacent balls is within the connection radius $r_j$. On the one hand, the radii $r'_j$ need to go to zero so that eventually, the trajectory remains in the $\delta$-neighborhood of $\gamma$ and therefore does not intersect any obstacles. On the other hand, one wants to ensure that for some $j$ \emph{each} of the balls $B(x_i, r'_j)$ contains a sample. One computes the probability $p_j$ of this event and then uses the Borel-Cantelli Lemma to conclude that, as long as $\sum_{j=1}^\infty (1-p_j) < \infty$, then the event is guaranteed to happen for infinitely many choices of $j$. The condition $r_j \geq \gamma (\frac{\log j}{j})^{\frac{1}{Q}}$ is then tailored to ensure that this happens (note that in the original proof, the dimension is only used to compute the volume of balls). One then concludes that for arbitrarily large choices of $j$, one has a trajectory $\gamma_j$ that follows $\gamma$. One concludes that PRM* is probabilistically complete. A refinement of the argument using Poissonization shows that the samples used to build $\gamma_j$ can be guaranteed to lie arbitrarily close to the centers of the balls so that one furthermore has $\ell(\gamma_j)\rightarrow \ell(\gamma)$. Thus, one gets trajectories that approximate $\gamma$ arbitrarily well in length. Thus, PRM* is asymptotically optimal. 

The argument for RRT* is essentially the same; see  \cite{solovey2020revisiting} for details.  The primary difference is that the samples $v_i$ in the balls $B(x_i, r'_j)$ must be produced in the right order so that RRT* is able to connect each $x_{i+1}^*$ to $x_i^*$ or to a lower-cost edge. In particular, the trajectory $\gamma_j$ constructed by connecting all of the $x_i^*$ in the sequence is either chosen when building the tree or a lower-cost alternate trajectory $\gamma'_j$ joins the starting point of the search with the point $x^*_{M_j}$. In either case, one concludes that $\lim_{n\rightarrow \infty} \ell(\gamma'_j)\leq \lim_{j\rightarrow \infty}  \ell(\gamma_j) = \ell(\gamma)$. The correct sampling order of the points $x^*_i$ is achieved by breaking up the time interval $[0, n]$ into $M_j$ approximately-equal intervals and requiring $x^*_i$ to be sampled in the $i^{th}$ time interval. Thus, the probability analysis for the sampling process takes place in the space $X\times \R$, of dimension $Q+1$ rather than $Q$, affecting the constraint on the connection radius sequence. One furthermore argues that step (2b) of the algorithm does not affect the long-term distribution of the points; indeed, the RRT part of the algorithm ensures that the vertex set is asymptotically dense, so the nearest connections become arbitrarily short after an initial exploratory period.
\end{proof}

\begin{remark}
One can remove step (2) from RRT* without losing the asymptotic optimality guarantees of Theorem \ref{thm:generalized} (although one \emph{would} lose the ability to produce RRT-type exploratory solutions in the short term). 
Indeed, the proof does not make use of this step, apart from needing to show that step (2b) does not affect the long-term distribution of points.
\end{remark}

\begin{remark}
If we assume that the free space $\Xfree$ is \emph{open}, the assumption of robust optimality is unnecessary. Optimal trajectories may not exist in this case, but any trajectory has strong $\delta$-clearance for some $\delta>0$.
\end{remark}

\subsection{Collaborative trajectory planning with convergence guarantees}
\label{subsec:collaborativegeneral}
We will now define what it means to combine multiple search problems into a single \emph{collaborative problem} and will prove:

\begin{thm}
\label{thm:joint}
Given individual search problems that satisfy Assumptions \ref{ass:basic}, any joint search problem corresponding to a parameter $1\leq p \leq \infty$ also satisfies Assumptions \ref{ass:basic}.
\end{thm}

Theorem \ref{thm:joint} relies on the following definition, including the use of a \emph{coupling parameter} $1\leq p\leq \infty$, which is motivated and explored in \ref{sec:applications}.
\begin{defi}
\label{defi:joint}
For $i=1, \ldots, n$, consider the \emph{individual} search problem with data  $(X_i, d_i, \mu_i)$, $(\Xsearch)_i$, $(\Xfree)_i$, $\Extend_i$, and $\random_i$.
A \emph{joint search problem} corresponding to a \emph{coupling parameter} $1\leq p\leq \infty$ is a search problem consisting of the following data:
\begin{enumerate}
    \item A joint state space $X=X_1\times \cdots \times X_n$, given by the Cartesian product of individual spaces and equipped with an $\ell^p$ combination $d$ of the metrics $d_i$ and the product measure $\mu_1 \times \cdots \times \mu_n$,
    \item The joint search space $\Xsearch=(\Xsearch)_1 \times \cdots \times (\Xsearch)_n$,
    \item Any choice of free space $\Xfree\subset \Xsearch$ containing the set $\{(x_1, \ldots, x_n)\in X \st \text{for all } i=1,\ldots, n, x_i\in (\Xfree)_i\}$,
    \item An $\ell^p$ combination $\Extend$ of the trajectories $\Extend_i$,
    \item A random variable $\random$ on $\Xsearch$ given by $\random=(\random_1, \ldots, \random_n)$.
    \item Any choice of the remaining data.
\end{enumerate}
\end{defi}
\begin{remark}
Generally, $\Xfree$ needs to be further restricted to avoid collisions between the systems, e.g., by removing some neighborhood of the set
$$\{(x_1, \ldots, x_n)\in X \st \text{ for some }i\neq j, x_i=x_j\}.$$
Note that in non-holonomic planning, only the location coordinates should be compared rather than the full pose. 
\end{remark}

We now elaborate on Definition \ref{defi:joint}. The products of spaces are Cartesian products. The product measure is the unique Borel measure such that for any Borel set of the form $A=A_1 \times \cdots \times A_n$ one has $\mu(A) = \mu_1(A_1) \cdots \mu_n(A_n)$. In particular, if the measures $\mu_i$ are Lebesgue measures, then $\mu$ is again a Lebesgue measure.

We combine the distances, and trajectories using the $\ell^p$ norm, which is a norm on $\R^n$ given by:
\begin{equation}
    \label{eqNorm}
\Norm{(v_1, \ldots, v_n)}_p =\begin{cases} 
      \left(\sum_{i=1}^n \norm{v_i}^p\right)^{1/p} & 1\leq p<\infty \\
      \max_{i=1}^n \norm{v_i} & p=\infty 
   \end{cases}.
\end{equation}
\begin{defi} For our $1\leq p\leq \infty$, we define:

The $\ell^p$ combination of distances $d_1, \ldots, d_n$ is given by:
$$d((x_1, \ldots, x_n), (x'_1, \ldots x'_n)) = \Norm{(d(x_1, x_1'), \ldots, d(x_n, x_n'))}_p$$
The $\ell^p$ combination of trajectories $\gamma_i: [0, T_i]\rightarrow X_i$ is the trajectory $G=(g_1, \ldots, g_n): [0, \Norm{(T_1, \ldots, T_n)}]\rightarrow X=X_1\times \cdots \times X_n$ given by:
$$g_i(t) = \gamma_i\left(t \frac{T_i}{\Norm{(T_1, \ldots, T_n)}_p}\right).$$
\end{defi}

\begin{remark}
The $\ell^p$ combination of the geodesics is defined such that the individual systems arrive at their destinations simultaneously. When $p=1$ or $p=\infty$ (but not for $1<p<\infty$), other choices of geodesics are also available. For $p=1$, one may, e.g., ask the systems to move one at a time. For $p=\infty$, one may, e.g., allow ``faster'' individual systems to arrive at the destination and wait for ``slower'' ones to complete their maneuvers.  See \ref{sec:applications} for more details.
\end{remark}

\begin{lemma}
\label{lemma:multitrajectory}
The $\ell^p$ combination of unit-speed geodesics $\gamma_1, \ldots,\gamma_n$ with time intervals $[0, T_i]$ is a geodesic.
\begin{proof}
Using the definition $d$ and fact that each $\gamma_i$ is a unit-speed geodesic, we obtain
\begin{align*}
d_p(&G(s), G(t))= \\&=\Norm{\left(d(\gamma_i(s \frac{T_i}{\Norm{T_1, \ldots, T_n}}), \gamma_i(t \frac{T_i}{\Norm{T_1, \ldots, T_n}})\right)_{i=1}^n}_p \\&= \Norm{((t-s) \frac{T_i}{\Norm{T_1, \ldots, T_n}})_{i=1}^n}_p=t-s,
\end{align*}
as desired.
\end{proof}
\end{lemma}

\begin{lemma}
If each metric measure space $(X_i, d_i, \mu_i)$ is $Q_i$-Ahlfors-regular on small scales, then the metric measure space $(X, d)$ is $Q$-Ahlfors-regular on small scales for $Q=\sum_{i=1}^n Q_i$.
\begin{proof}
If $p=\infty$, then a metric ball in $X$ is of the form 
$$B((x_1, \ldots, x_n), r) = B(x_1, r)\times \cdots \times B(x_n,r)$$
Because $\mu$ is a product measure, we have 
$$\mu(B((x_1, \ldots, x_n), r)) = \mu(B(x_1, r))\mu(B(x_n,r))$$
The result is then immediate from the fact that each $X_i$ is $Q_i$-Ahlfors-regular on small scales.

For other values of $p$, one has the standard inequality
$$\Norm{v}_\infty \leq \Norm{v}_p \leq \Norm{v}_1 \leq n \Norm{v}_\infty$$
which implies the corresponding fact for distances:
$$d_{\ell^\infty}(x,x') \leq d_{\ell^p}(x,x') \leq n d_{\ell^\infty}(x,x').$$
Combining this with the case $p=\infty$ completes the proof.
\end{proof}
\end{lemma}

\section{Applications}
\label{sec:applications}
We now consider concrete applications of our methods, focusing on illustrating the approach.
We first consider coupled one-dimensional systems to clarify the meaning of the coupling parameter $1\leq p\leq \infty$. Next, we demonstrate trajectory planning for individual and collaborative Reeds-Shepp vehicles, for which Theorem \ref{thm:generalized} provides the first guarantee of convergence (cf.~\cite{karaman_sampling_2013}, which discusses closely-linked sub-Riemannian geometries).
Lastly, we demonstrate trajectory-finding in the fractal Sierpinski Gasket geometry, which requires us to use highly non-Euclidean distances, measures, and sampling methods.

\subsection{Coupled robotics systems: a simple example}
\label{sec:coupledsimpleexample}

\begin{figure}[ht]
\centering
\includegraphics[width=.4\columnwidth]{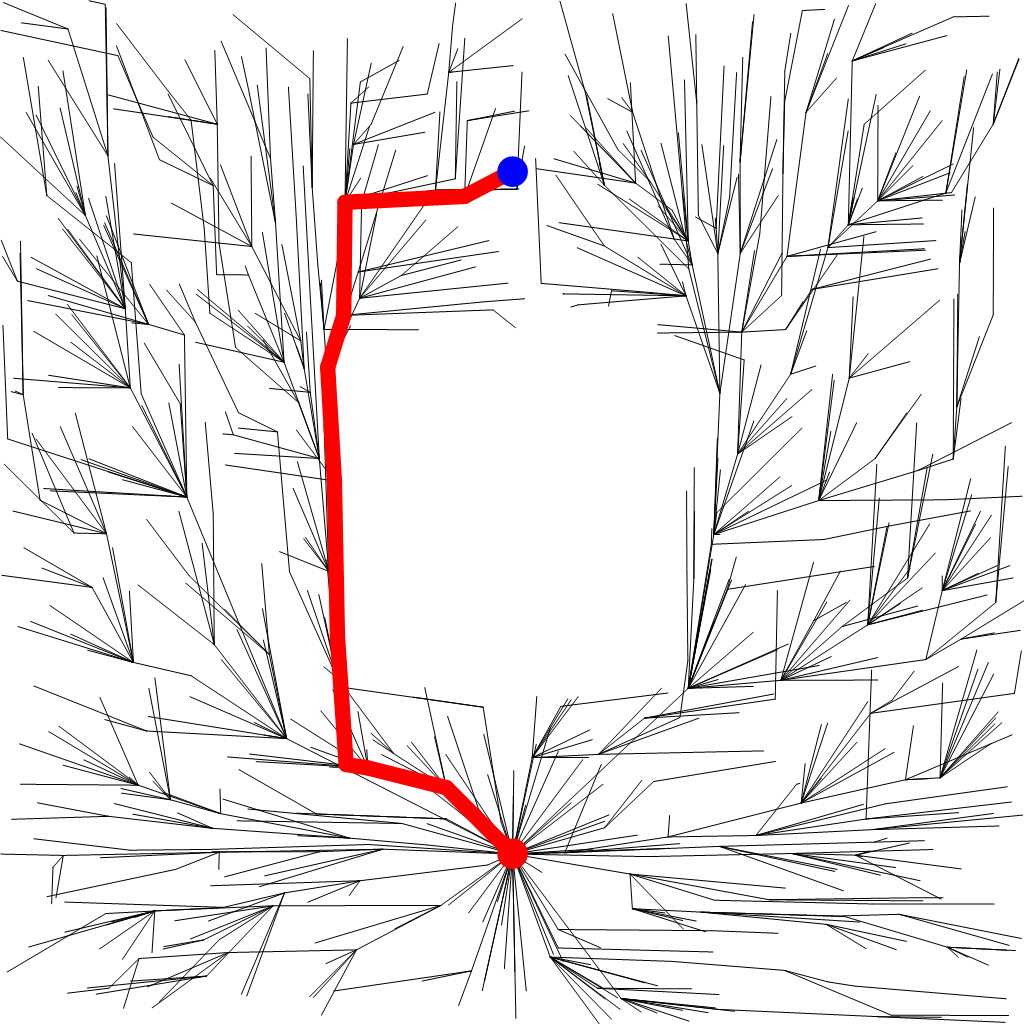}\\
\includegraphics[width=.4\columnwidth]{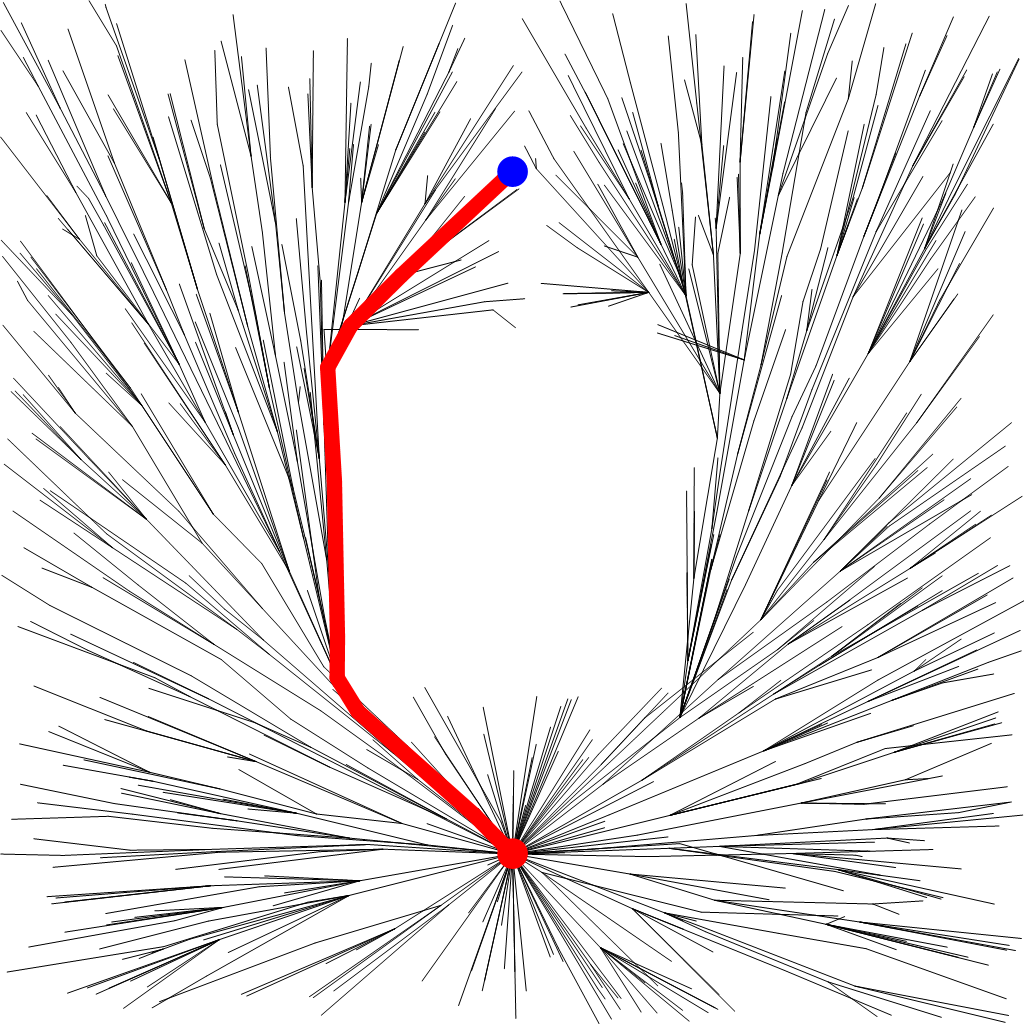}\\
\includegraphics[width=.4\columnwidth]{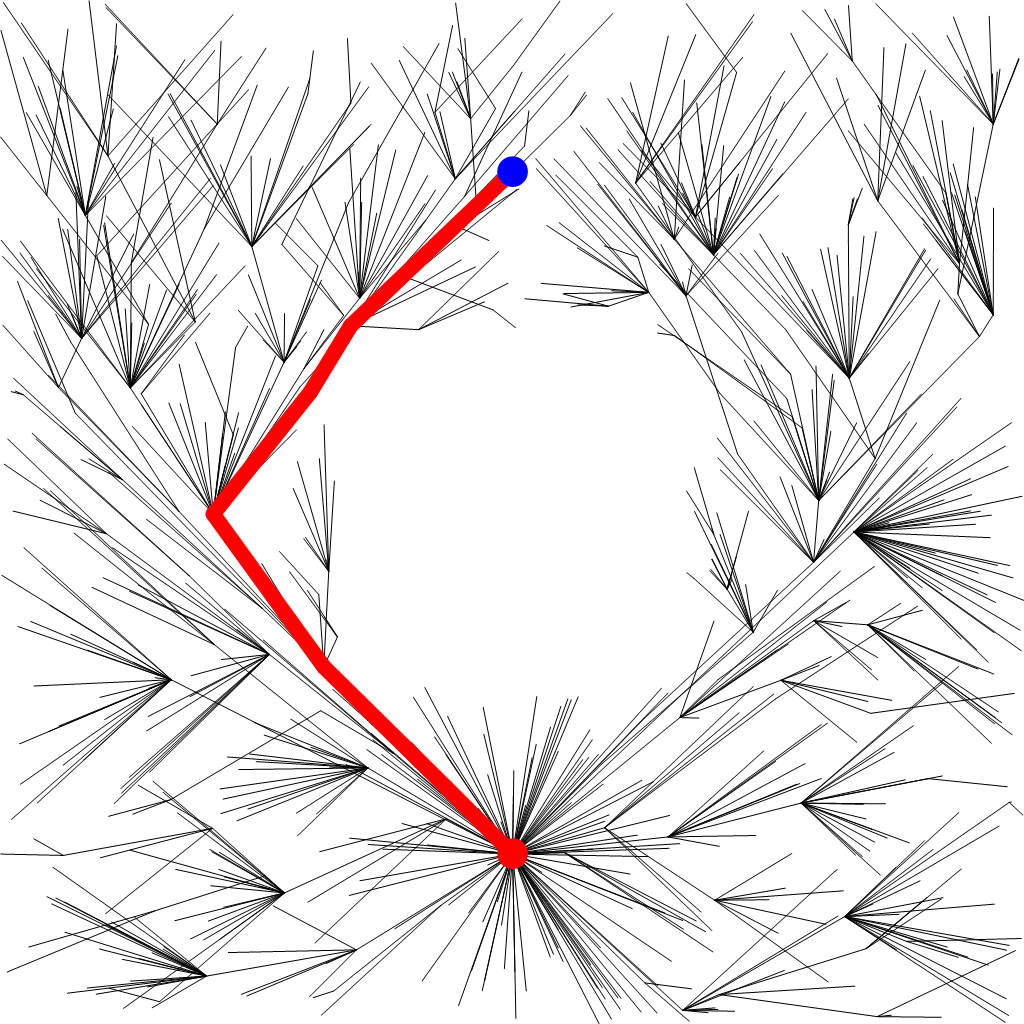}
\caption{trajectory planning in $\R^2$ with respect to the $\ell^1$, $\ell^2$, and $\ell^\infty$ metrics (top, middle, bottom, respectively)  produces markedly different trajectories. Here, trajectory planning occurs in the square $[0,3]\times [0,3]$  while avoiding the middle square $[1,2]\times[1,2]$. The tree produced by $RRT^*$ is shown, along with the best trajectory from $(1.5,0.5)$ to $(1.5,2.5)$. Each model uses the same 1000 samples.}
\label{fig:normcomparison}
\end{figure}

\begin{figure}[ht]
\includegraphics[width=.2\textwidth]{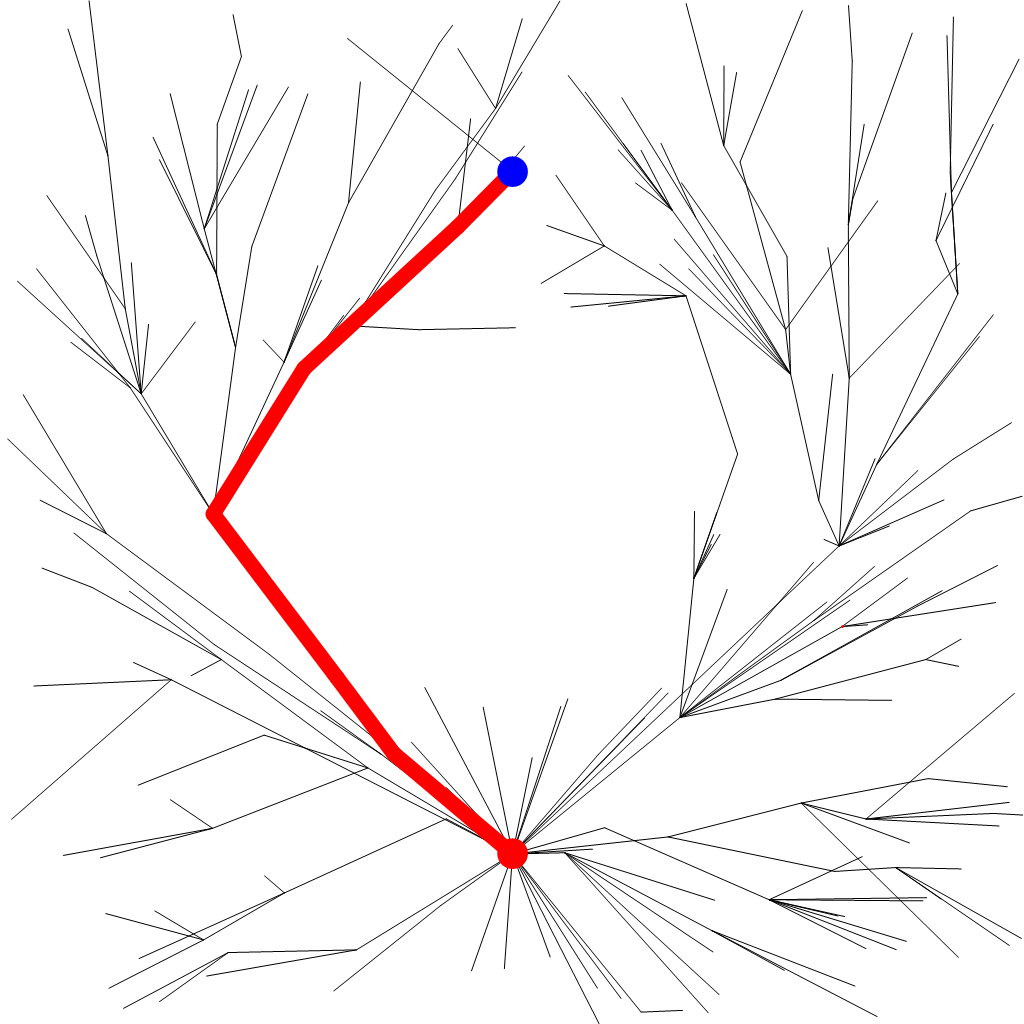}\hfill{}
\includegraphics[width=.2\textwidth]{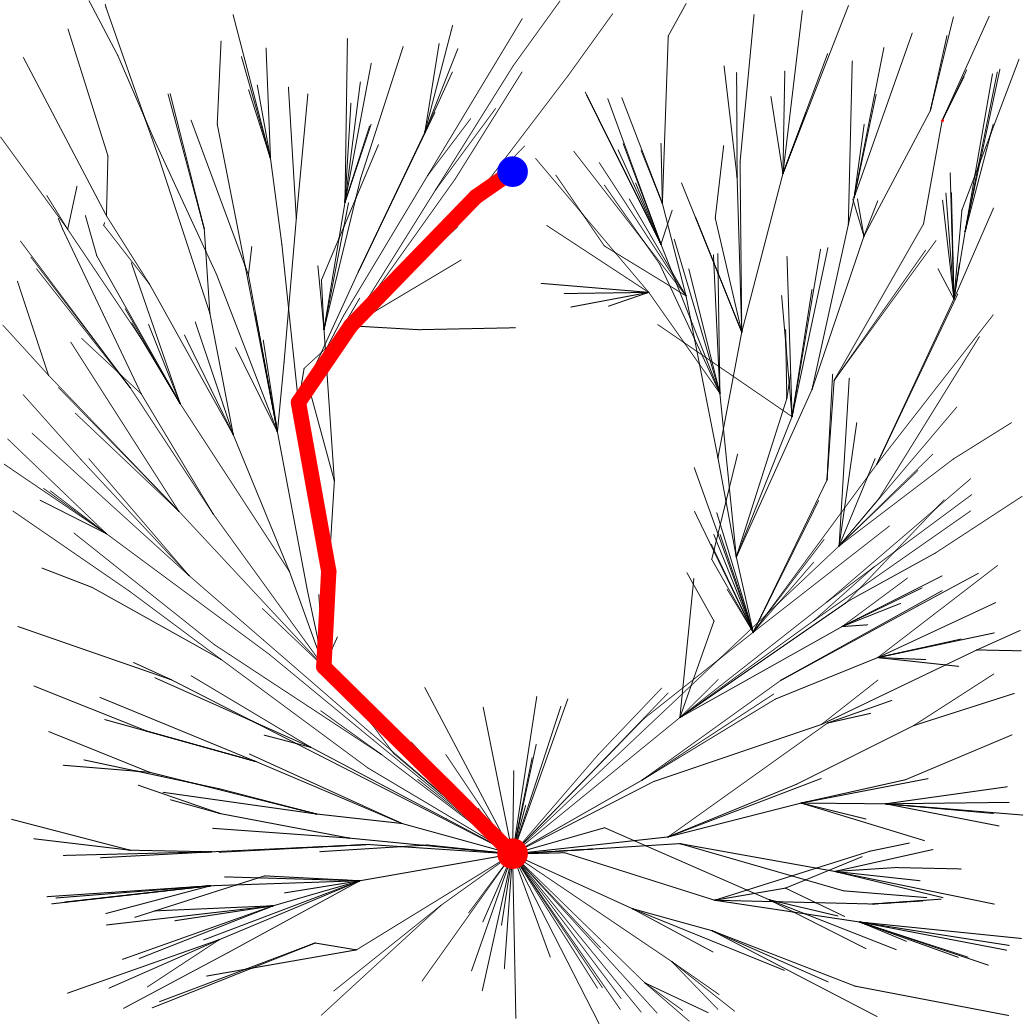}\hfill{}
\includegraphics[width=.2\textwidth]{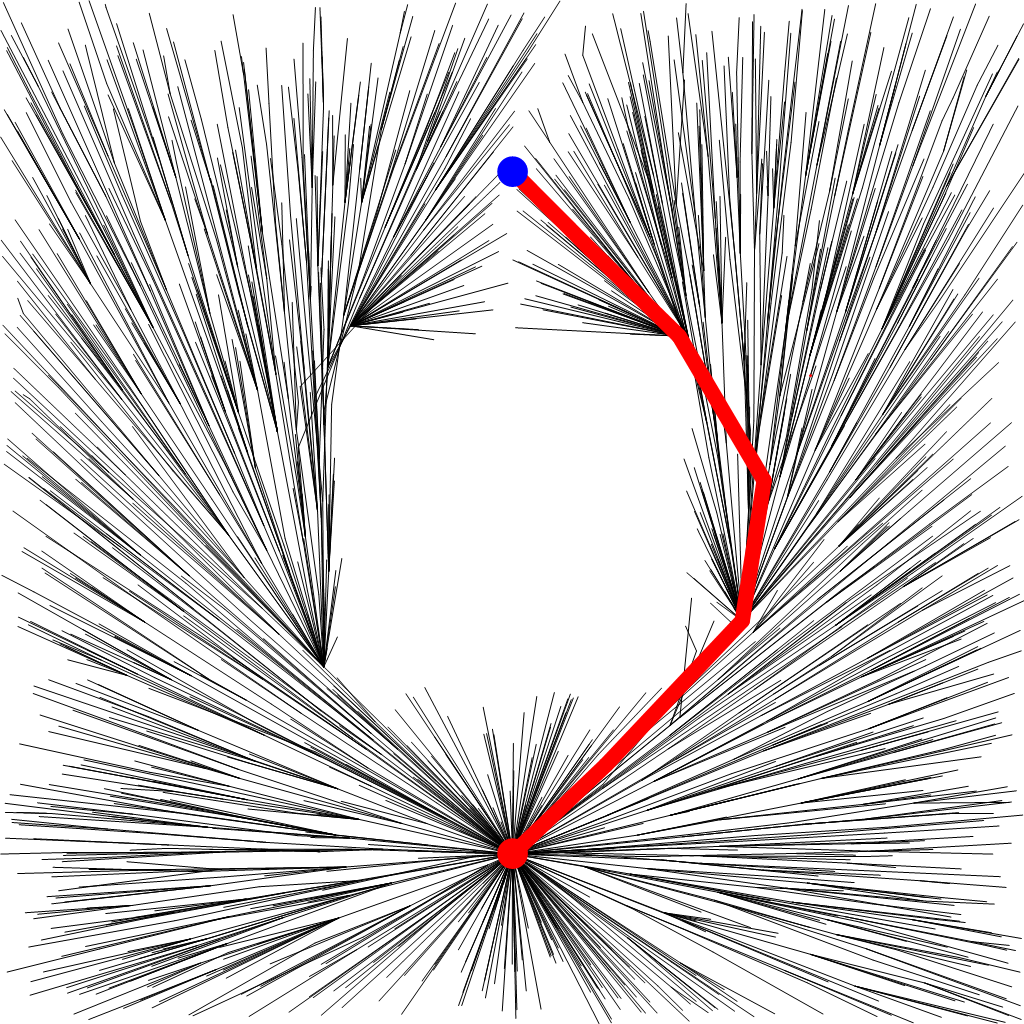}\hfill{}
\includegraphics[width=.2\textwidth]{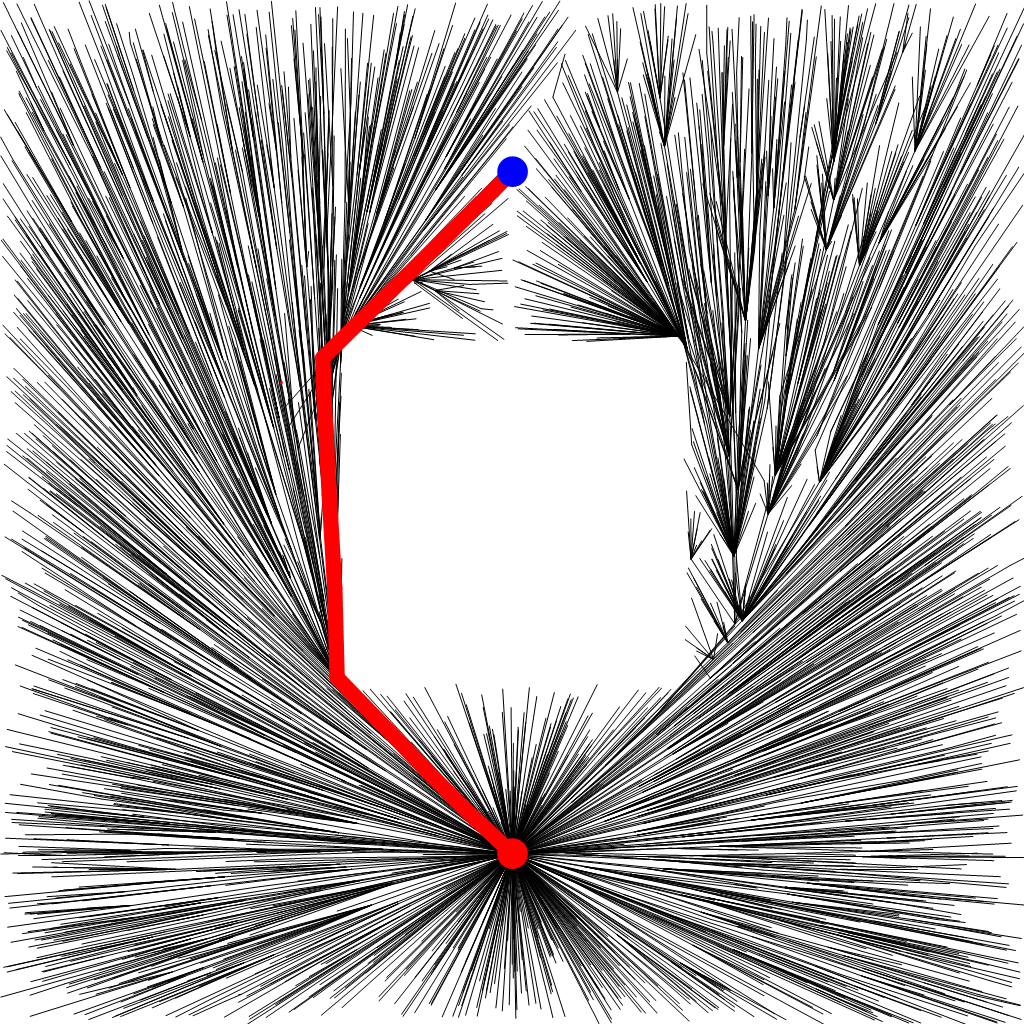}
\caption{Evolution of the RRT* tree for the $\ell^{p}$ metric with $p=10$. The high value of $p$ causes $\ell^\infty$-like behavior at first, and eventual convergence to a streamlined solution.}
\label{fig:lten}
\end{figure}
Consider a robotic system consisting of two robots $R_x$ and $R_y$ moving along independent linear rails, whose positions are given by variables $x$ and $y$, respectively. 

Let us explicitly write down a search problem for one of them, say, $R_x$. We work with the metric-measure space $(\R, d(x,x')=\norm{x-x'}, \mathcal L^1)$ where $\mathcal L^1$ is Lebesgue measure. This is, of course, a space that is $Q$-Ahlfors-regular on small scales with $Q=1$. We will work with $\Xsearch=[0,3]$ and $\Xfree=\Xsearch$. We take $\Extend(x,x')(t)=x+t\frac{x-x'}{\norm{x-x'}}$, for $t\in [0,\norm{x-x'}]$, and $\random$ a uniformly-distributed random variable on the interval $[0,3]$.

We build a joint planning task as described in Section \ref{subsec:collaborativegeneral}, further restricting $\Xfree$ to exclude the square $[1,2]\times[1,2]$. The configuration space for the joint system is then $\R^2$ equipped with a metric $d((x_0,y_0), (x_1,y_1))$ and Lebesgue measure $\mathcal L^2$. Depending on the coupling parameter $p$, we obtain three reasonable choices: the $\ell^1$ Manhattan metric given by $\norm{x_0-x_1}+\norm{y_0-y_1}$,  the familiar $\ell^2$ Euclidean distance, and the $\ell^\infty$ maximum norm given by $\max(\norm{x_0-x_1}, \norm{y_0-y_1})$. The $\ell^1$ metric can be interpreted as computing the total energy expenditure, while the $\ell^\infty$ metric computes the fastest transition time. The Euclidean metric does not have a natural interpretation but provides a middle point between $\ell^1$ and $\ell^\infty$ among the $\ell^p$ norms given by $\sqrt[p]{\norm{x_0-x_1}^p+\norm{y_0-y_1}^p}$ for $1\leq p<\infty$.
More generally, the coupling parameter $p$ calibrates the trade-off between total trajectory length minimization (when $p=1$) and total time minimization (when $p=\infty$).

Trajectory planning with respect to the $\ell^1$, $\ell^2$, and $\ell^\infty$ metrics produces markedly different results (Figure~\ref{fig:normcomparison}). The $\ell^1$ metric imposes no penalty for traveling along the coordinates one at a time since the corresponding cost is additive and therefore results in the coordinates mostly changing in sequence. For multi-robot systems, this corresponds to the $\ell^1$ metric inadvertently incentivizing excessive sequential vehicle motion. Conversely, the $\ell^\infty$ metric inadvertently incentivizes excessive joint motion, since the slower robotic system can move without imposing an additional total cost. This results in unnecessary motion, as seen in Figure~\ref{fig:normcomparison}, where the optimal trajectory moves further left along the $x$-axis than is strictly necessary, since such motion is not penalized by the $\ell^\infty$ metric. The Euclidean $\ell^2$ metric provides a middle ground between the two options: both excessive waiting and excessive joint motion are penalized.

Other choices of $\ell^p$ distances, with $1< p < \infty$, can also be used, with low values of $p$ favoring joint motion and high values of $p$ favoring individual motion. 
Because free-space optimal trajectories for $\ell^p$ distances with $1< p < \infty$ are straight lines, in \emph{simple cases} the long-term behavior of trajectory planners will resemble the Euclidean solution  Figure~\ref{fig:lten}. However, in complex scenarios, different choices of $p$ will lead to different trajectories.

\begin{figure}[ht]
\centering
\includegraphics[width=.6\columnwidth]{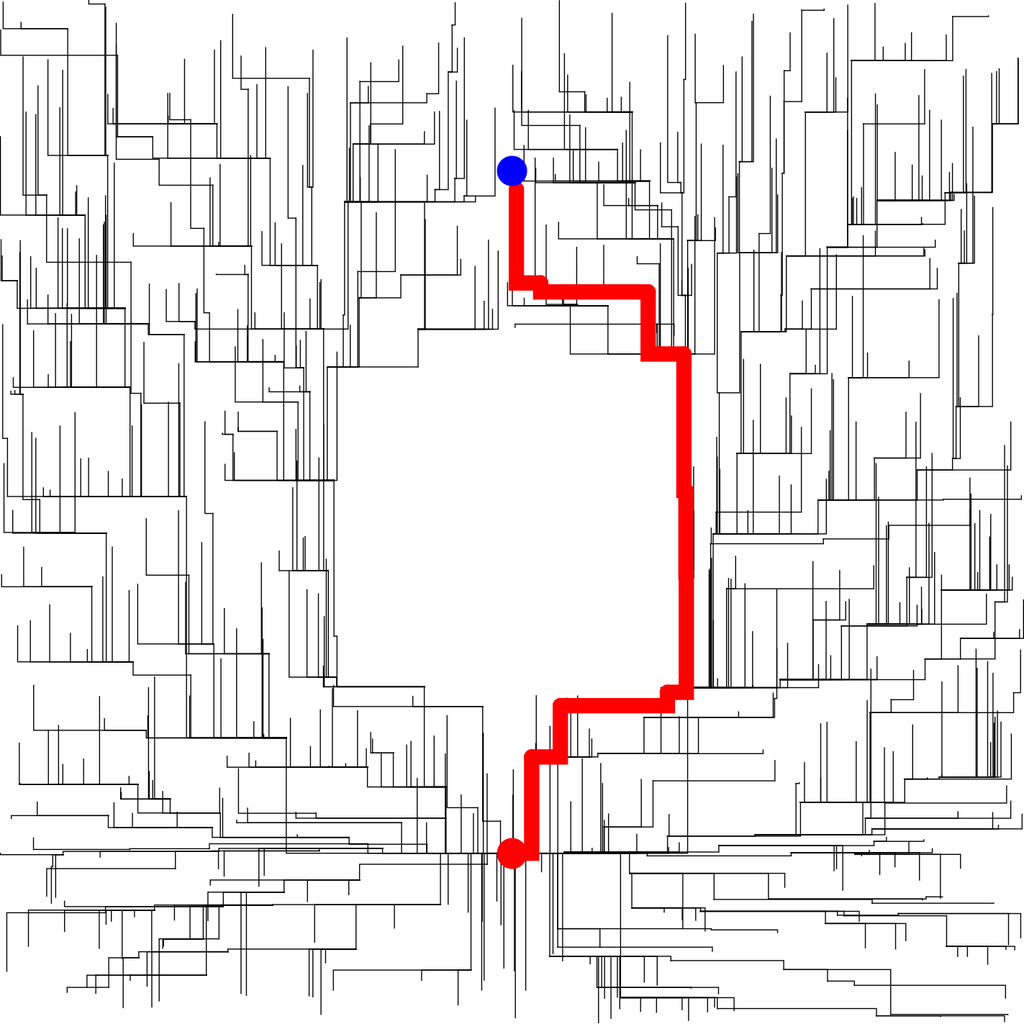}
\caption{An implementation of RRT* using a Manhattan \Extend{} function is asymptotically optimal with respect to the  $\ell^1$ cost (shown), but not with respect to the Euclidean cost, since it produces staircase trajectories whose Euclidean length corresponds to the \emph{Manhattan} distance.}
\label{fig:manhattan}
\end{figure}

Lastly, we demonstrate the fact that the choice of changing the \Extend{} function, by working with   $\R^2$ with the $\ell^1$ metric and the \emph{Manhattan trajectories}, which adjust each coordinate one at a time. Asymptotically-optimal $\ell^1$ trajectories are again recovered, as in Figure~\ref{fig:manhattan}. 
\begin{remark}
This example additionally demonstrates the reason we do not consider auxiliary cost functions in our framework. Namely, if we were interested in identifying Euclidean shortest trajectories using a Manhattan $\Extend$ function, we would fail. This is, perhaps, surprising given that the Manhattan metric and Euclidean metric are related by the \emph{bi-Lipschitz} inequality
$$d_{\ell^2}(p,q) \leq d_{\ell^1}(p,q) \leq \sqrt{2} d_{\ell^2}(p,q),$$
which preserves many geometric notions.
It is, therefore, unclear under what conditions on a cost make it compatible with a search framework.
\end{remark}

\subsection{Coupled Reeds-Shepp Vehicles}
\label{sec:fleets}
Let us describe a more involved example, using car-like robots. We prove in Section \ref{subsec:modelingRS} that a single-vehicle search problem for the Reeds-Shepp vehicle satisfies Assumptions \ref{ass:basic}, then discuss the joint planning space in further detail in Section \ref{sec:modelingfleet}, and demonstrate joint trajectory planning for two and three vehicles in \ref{subsec:multipleRSresults}

\subsubsection{Reeds-Shepp car}
\label{subsec:modelingRS}
The Reeds-Shepp car model is the standard representation of a vehicle for trajectory planning problems in low-speed environments. It represents a vehicle using three coordinates: the planar location $(x,y)$ and a heading $\theta$. Motion  in response to the velocity action $v$ and steering action $u$ is modeled using the equations
\begin{align} \label{eq:reedssheppfirst}
&\dot{x}=v \cos \theta &\dot{y}=v \sin \theta\\
&\dot{\theta}=uv &|u|\leq \frac{|v|}{\rho},
\end{align}
where 
$\rho$ is the turning radius of the vehicle.
The Reeds-Shepp model enforces the $\norm{v}\leq 1$ constraint. According to Reeds-Shepp theory~\cite{reeds_optimal_1990,sussmann_shortest_1991}, optimal trajectories for the Reeds-Shepp vehicle are given by a combination of straight trajectories and maximal-angle turns with $\norm{v}=1$. 

\begin{remark}
One can furthermore impose the constraint $v\geq 0$, producing a \emph{Dubins} vehicle model. The Dubins vehicle is not short-time controllable (e.g.~backwards displacement is only possible via long trajectories), finite-length trajectories are not reversible, and its state space is not modeled by a metric space. We will, therefore, not consider it here.
\end{remark}

\begin{figure}
    \centering
    \includegraphics[width=0.45\textwidth]{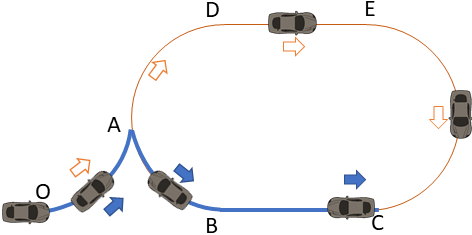}
    \caption{Optimal Reeds-Shepp (Blue line) and Dubins  (Orange line) trajectories.}
    \label{fig:RStrajectory}
\end{figure}

\begin{figure}[ht]
    \centering
    \includegraphics[height = 0.45\columnwidth]{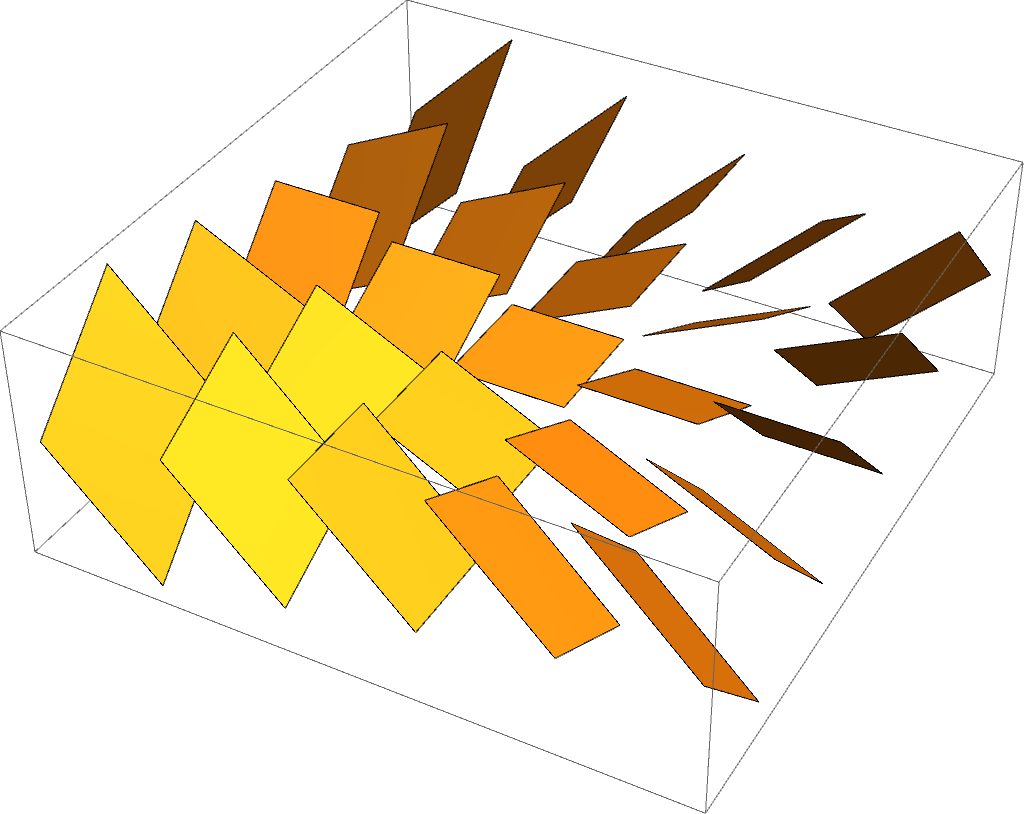}
    \caption{Sub-Finsler geometry encodes non-holonomic controls by specifying a linear subspace of controllable directions at each point.}
    \label{fig:SR_Distribution}
\end{figure}

The state space $\widetilde{RT}$ for the Reeds-Shepp vehicle can be given as a metric-measure space of a \emph{sub-Finsler manifold}, see \cite{sussmann_shortest_1991, jean_control_2014}. Specifically, $\widetilde{RT}$ is the space $\R^3=\{(x,y,\theta)\}$ equipped with the vector fields  $\xi(x,y,\theta) = ( \cos(\theta), \sin(\theta), 0)$ and $\eta(x,y,\theta)=(0,0,1)$. One interprets $\xi$ as motion in the forward direction and $\eta$ as a rotation. A trajectory in $\widetilde{RT}$ is then a \emph{controllable} trajectory (also known as \emph{permissible} or \emph{horizontal}) if $(\dot x(t), \dot y(t), \dot \theta(t)) \in \operatorname{span} (\xi, \eta)$.
The plane spanned by $\xi$ and $\eta$ changes depending on the basepoint (Figure~\ref{fig:SR_Distribution}), so that combining forward/backward motion and left/right turns unlocks sideways motion, as is familiar from parallel parking. Mathematically, this is encoded by the  Lie bracket of the two vector fields: the $[\xi, \eta] = (-\sin(\theta), \cos(\theta), 0)$ encodes sideways motion.

Given a controllable trajectory $\gamma(t)$, one can write $\dot \gamma = v \xi + u \eta$. Choosing a norm $\Norm{\cdot}$ on $\operatorname{span}(\xi, \eta)$ allows one to compute the length of $\gamma$ as $\ell(\gamma)=\int \Norm{\dot \gamma(t)}dt$. One shows that any two points in $\widetilde{RT}$ are connected by a finite-length pair of points, and defines a \emph{sub-Finsler} metric $d(p,q)$ on $\widetilde{RT}$, given by the infimum of lengths of controllable trajectories joining $p$ and $q$. If the norm $\Norm{\cdot}$ is induced by an inner product, then the metric is called \emph{sub-Riemannian}. For the Reeds-Shepp vehicle, taking $\Norm{(u,v)}=\max(\norm{u}, 2\pi \rho \norm{v})$ gives a sub-Finsler metric $d$ such $d(p,q)$ equals the shortest possible transition time for a Reeds-Shepp vehicle from pose $p$ to pose $q$. (Note that while the Reeds-Shepp model appears to prohibit turning in place, one can effectively do so by rapidly alternating the motion direction and turning angle.) One then shows that small balls $B((x,y,\theta),r)$ are approximated (see the Ball-Box Theorem in \cite{jean_control_2014}) by boxes
\begin{align*}\operatorname{Box}((x,y,\theta),r)=&\{(x,y,\theta)+u \xi+v \eta  + w [\xi, \eta]: \\&\norm{\xi}\leq r, \norm{u}\leq r, \norm{v}\leq r, \norm{w}\leq \sqrt{r}\}.
\end{align*}
That is, each ball contains a box of a similar radius and is, in turn, contained in a slightly larger box. One concludes that the notion of continuity on $(X,d)$ is the same as on $\R^3$ with the Euclidean metric. Additionally, if we take $\mu$ to be Lebesgue measure on $\R^3$ we have that $\mu(B((x,y,z),r))$ is approximately $r^4$ and conclude that $(X, d,\mu)$ is $Q$-Ahlfors-regular on small scales for $Q=4$.

In practice, one can compute the Reeds-Shepp distance by finding optimal straight-trajectory and radial-trajectory distances as in Figure~\ref{fig:RStrajectory} and computing the length of the resulting trajectory. For our illustrations, we use the implementation of this algorithm in the OMPL Library \cite{kavrakilab_OMPL_url}.

From the description of the underlying geometry, we conclude:
\begin{thm}
\label{thm:RSplanning}
Consider any search problem incorporating the following data:
\begin{enumerate}
    \item The space $\widetilde{RT}$ equipped with the Reeds-Shepp distance $d$ and Lebesgue measure $\mu$,
    \item Any $\Xfree\subset \Xsearch\subset X$ such that $0\leq \mu(\Xsearch)\leq \infty$,
    \item A function $\Extend$ that provides Reeds-Shepp geodesics,
    \item A function $\random$ that is uniformly distributed on $\Xfree$.
\end{enumerate}
Then the search problem satisfies Assumptions \ref{ass:basic} with $Q=4$. In particular, one obtains convergence guarantees for PRM*, RRT, and RRT* as in Theorem \ref{thm:generalized}, using $Q=4$.
\end{thm}

\subsubsection{Fleet of Reeds-Shepp cars}\label{sec:modelingfleet}

A search problem for a fleet of $n$ Reeds-Shepp vehicles can be immediately constructed as in Section \ref{sec:coupledsimpleexample}, with convergence guarantees coming from Theorems \ref{thm:generalized}, \ref{thm:joint}, and \ref{thm:RSplanning}. Let us describe a particular implementation of this joint search process more explicitly.

We represent a fleet of $n$ cars as a Cartesian product of individual vehicle configurations. That is, a fleet is represented by $3n$ coordinates $(x_1, y_1, \theta_1), \ldots,$ $(x_n, y_n, \theta_n)$, with each car $i$ satisfying the motion equations in \S\ref{subsec:modelingRS}. Correspondingly, we model the motion of the fleet as a single $3n$-dimensional joint motion function (or \emph{multitrajectory})
\begin{equation} \label{eq:trajectoryP} P(t)=(x_1(t), y_1(t), \theta_1(t), \ldots, x_n(t), y_n(t), \theta_n(t)) \end{equation}
that can be decomposed, via projection, into trajectories 
\begin{align}
	p_i(t)&=(x_i(t), y_i(t), \theta_i(t))
\end{align}
for each car $i$. 

Conversely, suppose we are given starting and ending poses for the fleet:
\begin{align}
a =(x_1, y_1, \theta_1, \ldots, x_n, y_n, \theta_n),\\
b=(x'_1, y'_1, \theta'_1, \ldots, x'_n, y'_n, \theta'_n),
\end{align}
For each vehicle, let $T_1$ be the Reeds-Shepp distance between $(x_i, y_i, \theta_i)$ and $(x'_i, y'_i, \theta'_i)$ and $p_i^{RS}(t)$ the unit-speed Reeds-Shepp geodesic  joining these poses, with $t\in [0, T_i]$.  Fix a coupling parameter $1\leq p\leq \infty$. Then the distance $d$ on the joint planning space between $P$ and $P'$ is given by $\Norm{(T_i)}_p$, and we combine the trajectories into a  multi-trajectory $P(T)$ with $t\in [0, \Norm{(T_i)}_p]$ given by 
$$p_i(t) = p_i^{RS}\left(t \frac{T_i}{\Norm{(T_i)})_p}\right).$$
By Lemma \ref{lemma:multitrajectory}, $P(T)$ is then a unit-speed geodesic in the joint space.

\begin{remark}For $p=\infty$, the distance can be equivalently expressed as 
\begin{equation}
\label{eq:metric}    
d_\infty(a,b)=\min.~ \operatorname{Time}(P),
\end{equation}
where $\operatorname{Time}(P)$ is the time necessary to complete the multi-trajectory $P$ and the minimum is taken over all multi-trajectories $P$ joining $a$ to $b$, subject to the model's constraints. 
\end{remark}

\begin{remark}
Note also that at this stage of model development, we have not specified $\Xfree$, so that the vehicles do not detect any collisions, including collisions with each other, and may pass through each other. This allows us to obtain the optimal multi-trajectories $P(t)$ explicitly and will be rectified when we choose $\Xfree$ to avoid collisions below.
\end{remark}

Next, we equip our search space with a measure, search space, and random variable. The natural measure for the Reeds-Shepp vehicle happens to be the Lebesgue measure of $\R^3$, so the multi-vehicle space is again equipped with the Lebesgue measure on $\R^{3n}$. The resulting metric-measure-space is $Q$-Ahlfors-regular on small scales with dimension $Q=4n$, since the individual vehicle's configuration space is $Q_i$-Ahlfors-regular on small scales with $Q_i=4$.
We restrict our attention to a rectangular search region $\Xsearch=[0,W]^{3n}$, and give it a random variable $\random$ by combining $3n$-many copies of a uniform random variable on $[0,W]$.

In building $\Xfree$, we avoid three types of collisions: (i) inter-vehicle collisions, (ii) collisions by individual vehicles with environmental obstacles, and (iii) trajectories that allow an individual vehicle to leave the model's region (note that the trajectories are non-linear so this is not automatic). For collision-avoidance purposes, we model the vehicles as a union of two (Euclidean) disks of radius $r$ centered at the coordinates $F_i=(x_i, y_i)$ and $R_i=(x_i, y_i)-1.5r(\cos(\theta_i), \sin(\theta_i))$. Inter-vehicle collisions are prohibited by restricting the Euclidean distance between any pair of disks $i\neq j$:
\begin{align*}
d(F_i, F_j)\geq 2r, \hspace{.3in} &d(F_i, R_j)\geq 2r,\\
d(R_i, F_j)\geq 2r, \hspace{.3in} &d(R_i, R_j)\geq 2r.
\end{align*}

For purposes of the implementation, type 2 and type 3 constraints are replaced with a list of edges $E_j$ that may not be touched by any vehicle:
\begin{align*}
&d((x_i,y_i), E_j)\geq r,\\
&d((x_i-1.5r \cos(\theta_i),y_i-1.5 r\sin(\theta_i)), E_j)\geq r.
\end{align*}
where the Euclidean distance to the edge is computed as the minimum of the distance to the endpoints of $E_j$ and the distance to the perpendicular projection of $(x_i, y_i)$ to the line containing $E_j$.

\subsubsection{Simulations}\label{subsec:multipleRSresults}
We now demonstrate specific applications of the above search problem in the cases $n=2$ and $n=3$.

Note that precise joint motion planning for small numbers of vehicles is required even when planning for larger numbers of vehicles. In such settings, local conflicts in trajectories are resolved by locally  increasing the dimension of the problem in the relevant region. In the local \emph{sub-dimensional expansion} problem, a limited number of vehicles enter a box with specific poses and need to leave with desired poses while avoiding each other and other obstacles in the box. We assume that the problem cannot be simplified any further by reassigning the tasks or similar methods, and joint planning is needed.

\begin{figure}[ht]
	\centering
    \subfloat[$\ell^1$ multi-trajectory \label{fig:twocar_p1}]{
     \includegraphics[width = .45\columnwidth]{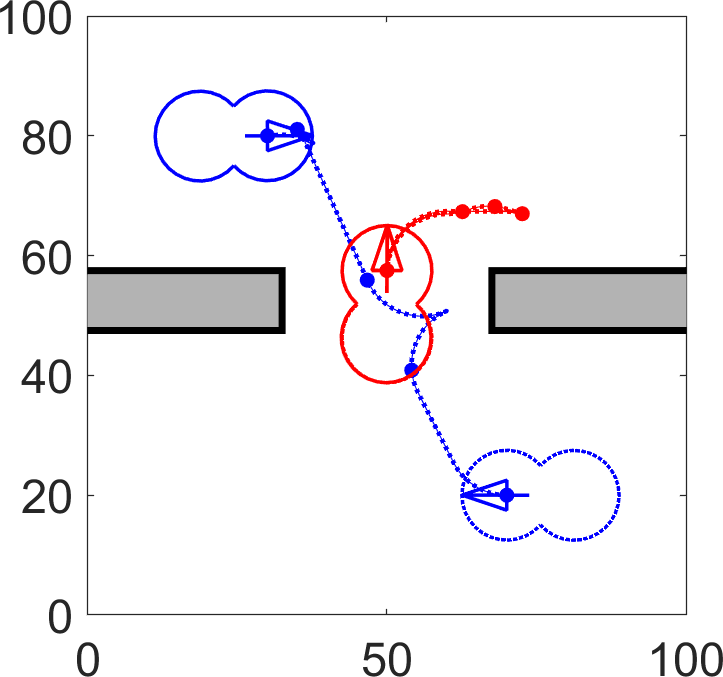}
    }
    \subfloat[$\ell^1$ speed profile \label{fig:speed_p1}]{
     \includegraphics[width = .4\columnwidth]{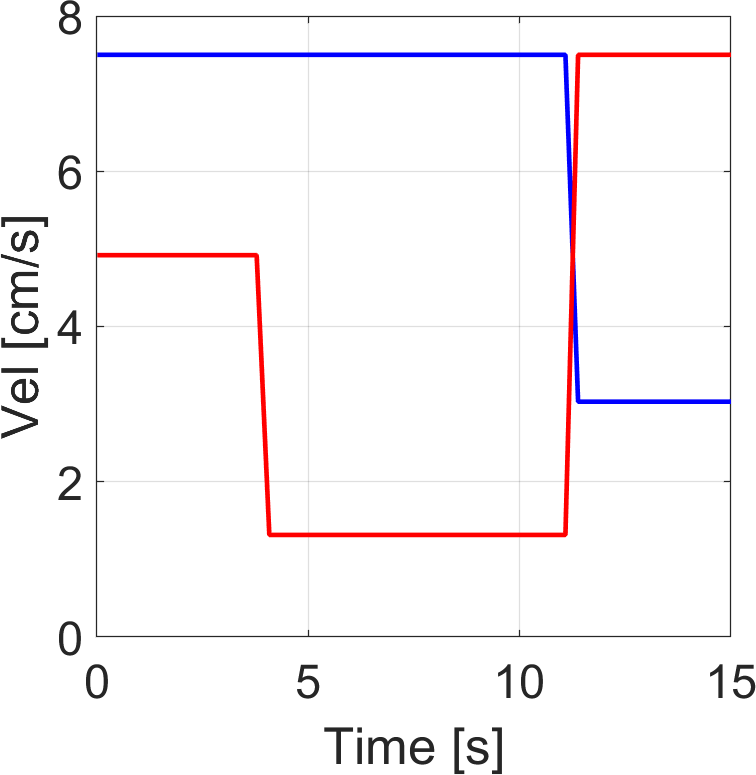}
    }
    \\
    \subfloat[$\ell^2$ multi-trajectory \label{fig:twocar_p2}]{
     \includegraphics[width = .45\columnwidth]{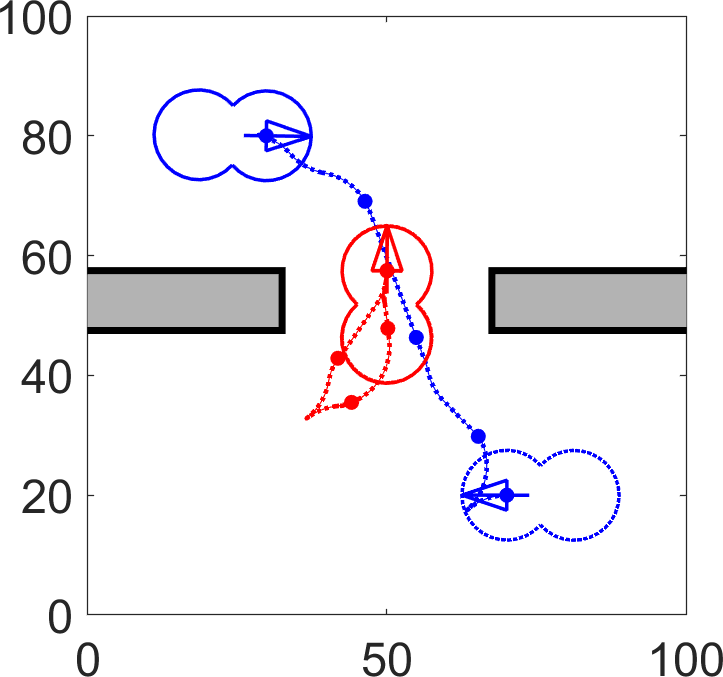}
    }
    \subfloat[$\ell^2$ speed profile \label{fig:speed_p2}]{
     \includegraphics[width = .4\columnwidth]{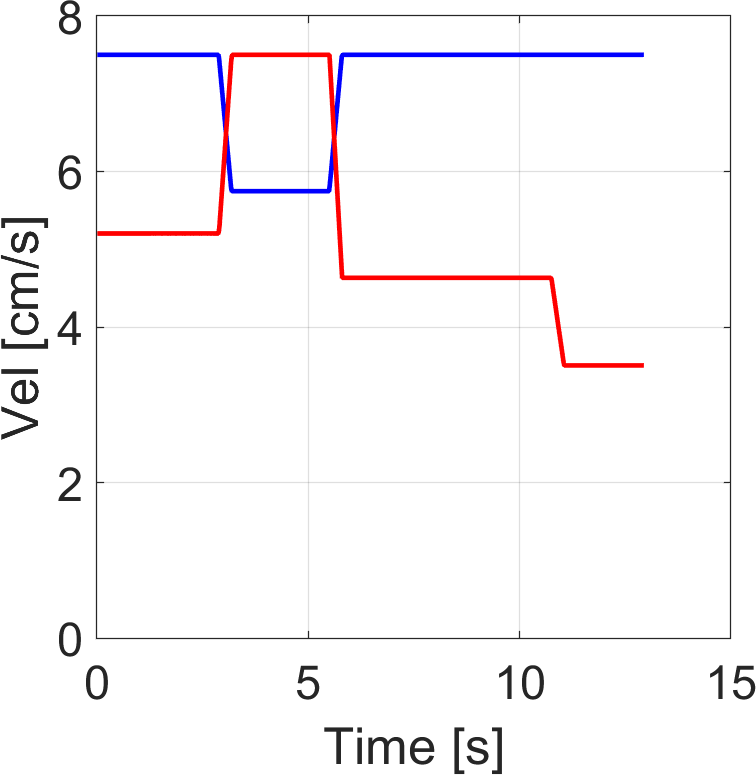}
    }
    \\
	\subfloat[$\ell^{\infty}$ multi-trajectory \label{fig:twocar_pinf}]{
       \includegraphics[width = .45\columnwidth]{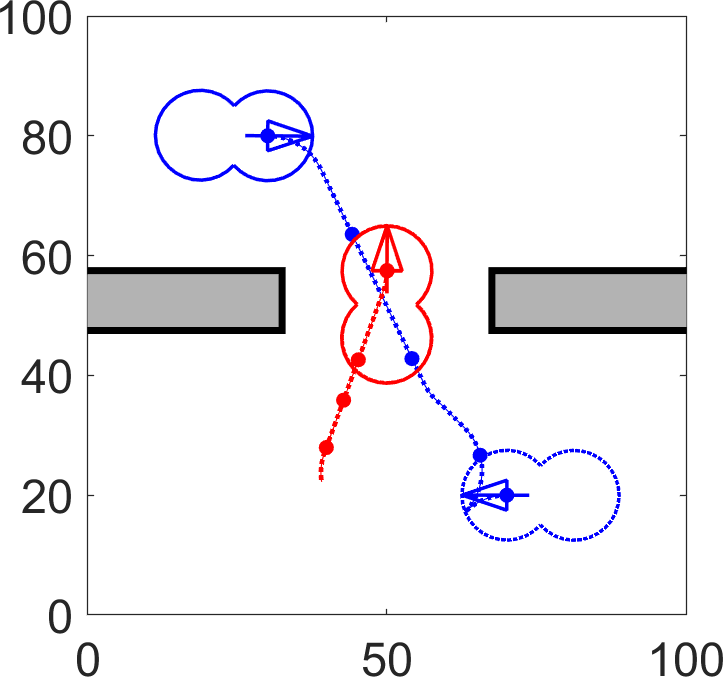}
     }
	\subfloat[$\ell^{\infty}$ speed profile\label{fig:speed_pinf}]{
       \includegraphics[width = .4\columnwidth]{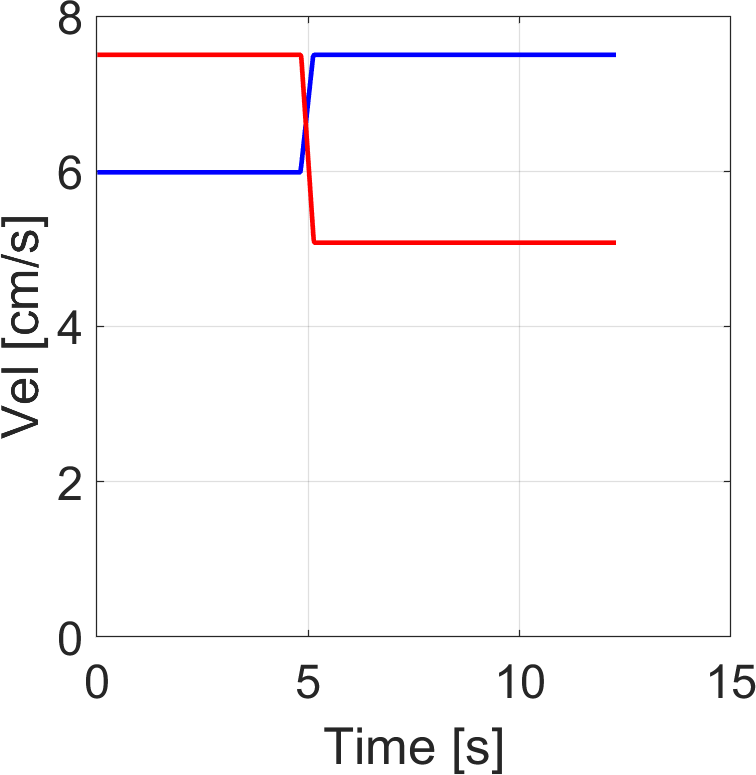}
     }
    \caption{Trajectory planning for two car-like robots in a 100x100 $cm^2$ region with obstacles using (a) $\ell^1$-norm, (c) $\ell^2$-norm, and (d) $\ell^{\infty}$-norm. We show the starting pose of each vehicle using dashed lines and arrows, and the final pose using solid lines and arrows. Static environmental obstacles are shown as gray boxes. The speed of each vehicle over time is shown in the corresponding figures (b), (d), (f).}
     \label{fig:twocar_norms}
\end{figure}

Consider first the scenario shown in Figure~\ref{fig:twocar_norms}, where two cars are navigating a small region with obstacles: the blue car is moving from the bottom-right corner of the region, with starting pose [70, 20, $\pi$], to the upper-left corner of the region, with ending pose [30, 80, 0]; while the red car is positioned in the middle of the region with pose [50, 57.5, $\pi/2$], and needs to return to the same pose. 

We perform trajectory planning with respect to coupling parameters $p=1$, $p=2$, and $p=\infty$, corresponding (see Sections \ref{subsec:collaborativegeneral}, \ref{sec:coupledsimpleexample}). The three planning tasks result in different multi-trajectories, shown in (a,c,e) of Figure~\ref{fig:twocar_norms}, each of which is near-optimal for the respective metric.

The distinction between the three multi-trajectories is apparent in the speed profiles\footnote{Note that the multi-trajectories are parametrized such that the faster car is always moving at the maximum allowed speed, which does \emph{not} correspond to a unit-speed parametrization in the $\ell^p$ metric.} seen in (b,d,f) in Figure~\ref{fig:twocar_norms}. We note that:
\begin{itemize}
    \item the $\ell^\infty$ planner provides the fastest multi-trajectory (12.3s), since the $\ell^\infty$ cost of the multi-trajectory corresponds exactly to the time required to traverse it; while the $\ell^2$ and $\ell^1$ multi-trajectories take longer to traverse (12.9s and 15s, respectively),
    \item the $\ell^1$ planner provides the smallest total amount of motion (152.4cm total), since the $\ell^1$ cost computes the total travel distance for both vehicles; while the $\ell^2$ and $\ell^\infty$ multi-trajectories have longer total travel distance (158.5cm and 158.6cm, respectively),
    \item the speeds of the two cars are highly matched in the $\ell^\infty$ case, which prioritizes joint motion, and is highly varying in the $\ell^1$ case, which prioritizes individual motion; with $\ell^2$ motion providing a middle ground.
\end{itemize}
The simulations thus illustrate the fact that the $p$ value in the  $\ell^p$ metric provides a way to choose the extent to calibrate the planner to the designer's preference in the tradeoff between fastest-motion or shortest-total-motion multi-trajectories.

\begin{figure}[ht]
	\centering
    \subfloat[Initial pose, $t=0$ \label{fig:threecar_f1}]{%
     \includegraphics[width = .47\columnwidth]{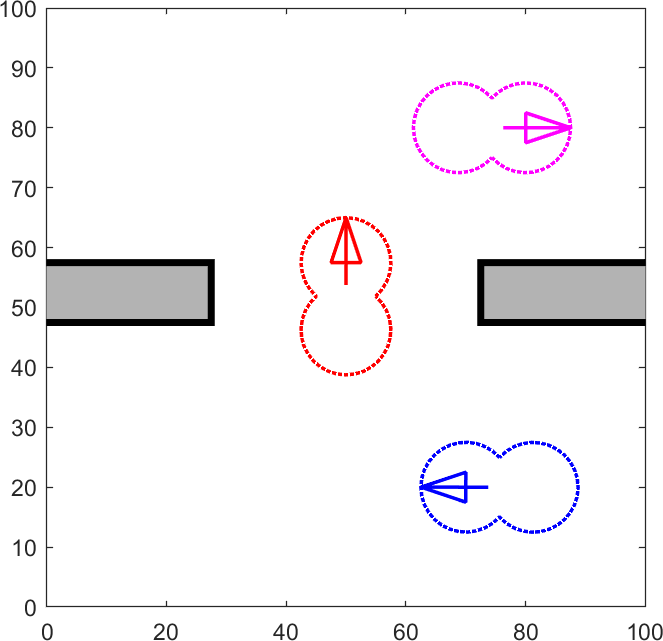}
    }
    \subfloat[Intermediate pose, $t=2s$ \label{fig:threecar_f2}]{%
     \includegraphics[width = .47\columnwidth]{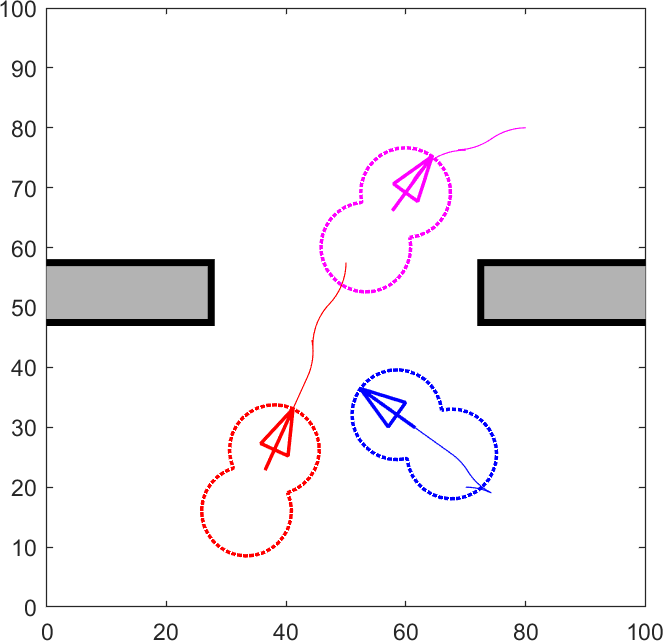}
    }\\
	\subfloat[Intermediate pose, $t=5s$\label{fig:threecar_f3}]{%
       \includegraphics[width = .47\columnwidth]{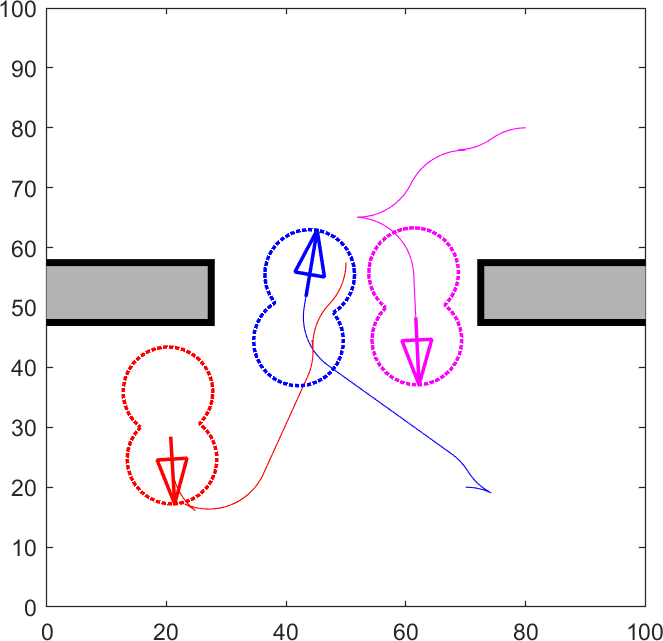}
     }
    \subfloat[Final pose, $t=10s$\label{fig:threecar_f4}]{%
       \includegraphics[width = .47\columnwidth]{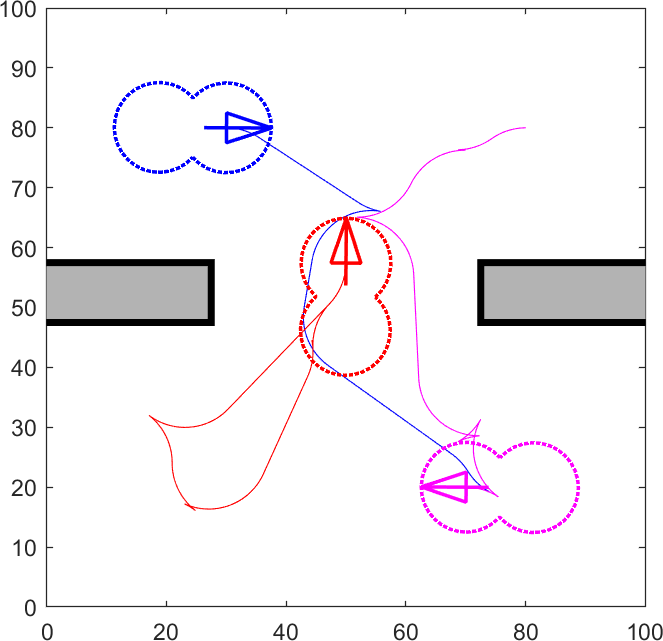}
     }
     \caption{Trajectory planning for three car-like robots in a 100x100 $cm^2$ area. The red robot returns to its initial pose, while the blue and magenta robots switch sides.}
     \label{fig:threecar}
   \end{figure}

Next, Figure~\ref{fig:threecar} shows the simulation results for three Reeds-Shepp cars maneuvering in an environment with polygonal obstacles with an $\ell^2$ cost function, which provides a compromise between the shortest-time and shortest-total-length options for multi-trajectory planning. 

In this example, the blue and magenta cars move from initial poses of $[70,20,\pi]$ and $[80,80,0]$, as shown in Figure~\ref{fig:threecar_f1}, to their destinations at $[30,80,0]$ and $[70,20,\pi]$, respectively, while the red gatekeeper car returns to its original pose $[50, 57.5,\pi/2]$ as shown in Figure~\ref{fig:threecar_f4}. Static environmental obstacles are shown as gray boxes.    
   
We show two important intermediate poses for the fleet. First, in Figure~\ref{fig:threecar_f2}, we see that the blue and magenta cars have changed direction and started to move toward the center of the region, while the red car has started to make space for their trajectories. Next, in Figure~\ref{fig:threecar_f3}, the red car is out of the way of the blue and red cars, and the blue and magenta cars move through the opened passage at the same time. Finally, the blue and magenta cars reach their destinations in Figure~\ref{fig:threecar_f4}, and the red car returns to its initial pose at the center of the region.

\subsection{Fractal trajectory Planning}
\newcommand{\Sierp}{\mathcal S}
\newcommand{\corner}{\operatorname{corner}}
Our assumptions, while motivated by vehicle trajectory planning, apply to a wide range of scenarios. We illustrate this by implementing RRT* in the \emph{Sierpinski gasket} $\Sierp$, a self-similar fractal set in $\R^2$ of Hausdorff dimension $Q=\frac{\log 3}{\log 2}$. 

We start by describing the Sierpinski gasket, making use of the notation and results of \cite{MR4261663}. A common description of $\Sierp$ is as follows: start with a solid equilateral triangle, remove the central triangle defined by the mid-points of the sides, and repeat the process iteratively with the remaining three triangles. To specify a point $p\in \Sierp$, we can specify a sequence of triangles that it resides in, e.g., the bottom-left corner of the top triangle is given by $p=(0,1,1,\ldots)$ where the $0$ indicates that it is in the top triangle, and the 1s indicate that under further subdivisions $p$ remains in the left triangle. Here, we use the convention that $0$ is the top triangle, $1$ is the left one, and $2$ is the right one. Note that infinite sequences correspond to points, while finite sequences can be interpreted as ``triangular'' subsets of $\Sierp$. As with real numbers, a point may have multiple descriptions, e.g., $(0,\overline 1)=(1,\overline 0)$. We can convert a digit sequence $(a_i)_{i=1}^\infty$ into Cartesian coordinates by taking $(x,y)=\sum_{i=1}^\infty 2^{-i}\corner(a_i)$ where $\corner(0)=(1/2, \sqrt{3}/2)$, $\corner(1)=(0,0)$, and $\corner(2)=(1,0)$.

Distances in $\Sierp$ are computed abstractly by taking $d(p,q)=\inf \ell(\gamma)$ where the infimum is taken over all finite-length trajectories $\gamma \subset \Sierp \subset \R^2$ joining $p$ and $q$. One can construct optimal trajectories explicitly as follows. Start by zooming in and orienting $\Sierp$ such that $p$ is in the left triangle and $q$ is in the right triangle, i.e.~after normalization we have $p=(1, p_2, p_3, \ldots)$ and $q=(2, q_2, q_3, \ldots)$. While there are, in some cases, as many as five optimal trajectories between $p$ and $q$, we construct a specific one as follows. One possibility is that an optimal trajectory passes through the mid-point indexed by $(1,\overline{2})=(2,\overline{1})$. From here, it goes left towards $p$ through the vertices $(1,p_2,\ldots, p_i, \overline{2})$, and to the right towards $q$ through the vertices $(2, q_2, \ldots, q_i,\overline{1})$. Another possibility is that an optimal trajectory passes through the higher vertices $(1,\overline 0)$ and $(2, \overline 0)$, in which case we connect $p$ to $(1,\overline 0)$ as before by replacing digits, then connect $(1,\overline 0)$ to $(2,\overline 0)$, and then interpolate towards $q$ as before. In our implementation, we compute both of these trajectories and return the shorter one.

A random point in $\Sierp$ can be selected by specifying a random infinite digit sequence, at each stage choosing uniformly between $0, 1, 2$. The resulting random variable is uniformly distributed with the measure $\mu$ that assigns each triangle of depth size $3^{-n}$. Equivalently, we can select a random point in the interval $[0,1]$, extract its base-3 digits, and reinterpret the sequence of digits as a point in $\Sierp$. 

The resulting metric-measure space is $Q$-Ahlfors-regular at small scales with $Q=\frac{\log 3}{\log 2}$, which can be seen from the self-similarity of the space. The identification of any of the three first-level triangles and the full space rescales the measure by a factor $3$ (since it is made up of three identical triangles) and distances by a factor $2$ (with the right choice of coordinates, the mapping is given by $(x,y)\mapsto (2x,2y)$ which doubles lengths).

\begin{figure}[ht]
    \centering
    \includegraphics[width=0.6\columnwidth]{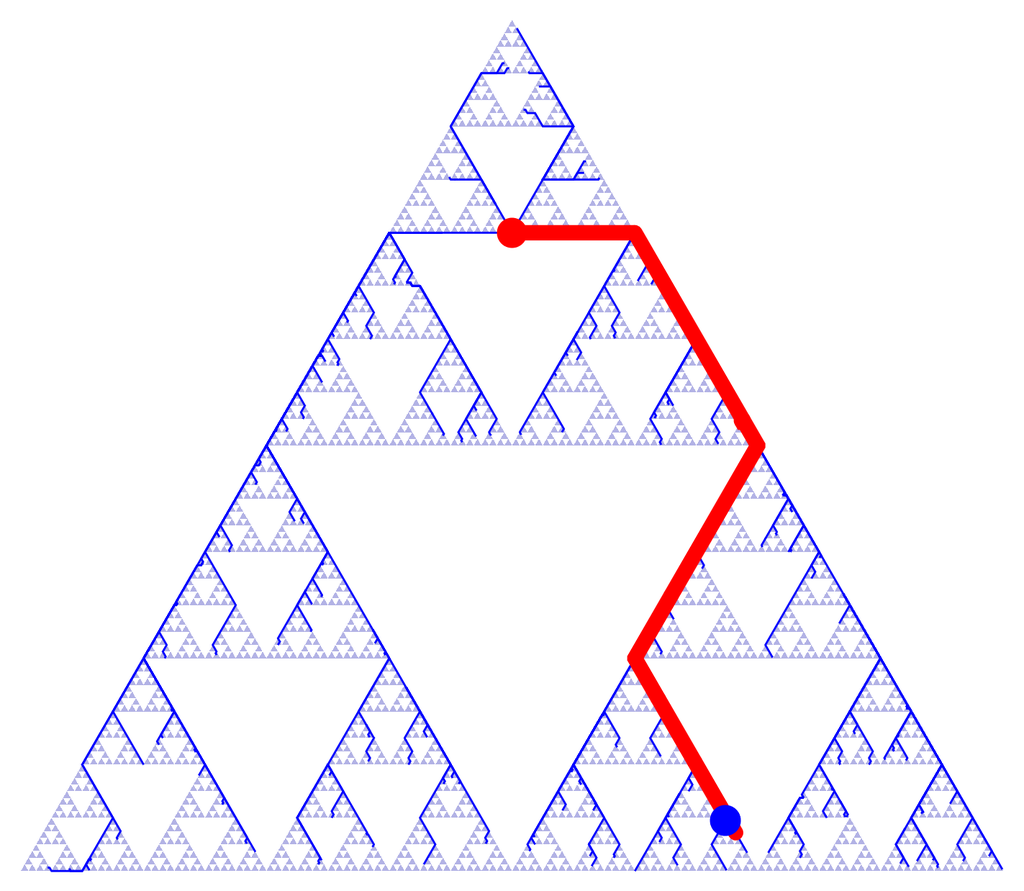}
    
    \includegraphics[width=0.6\columnwidth]{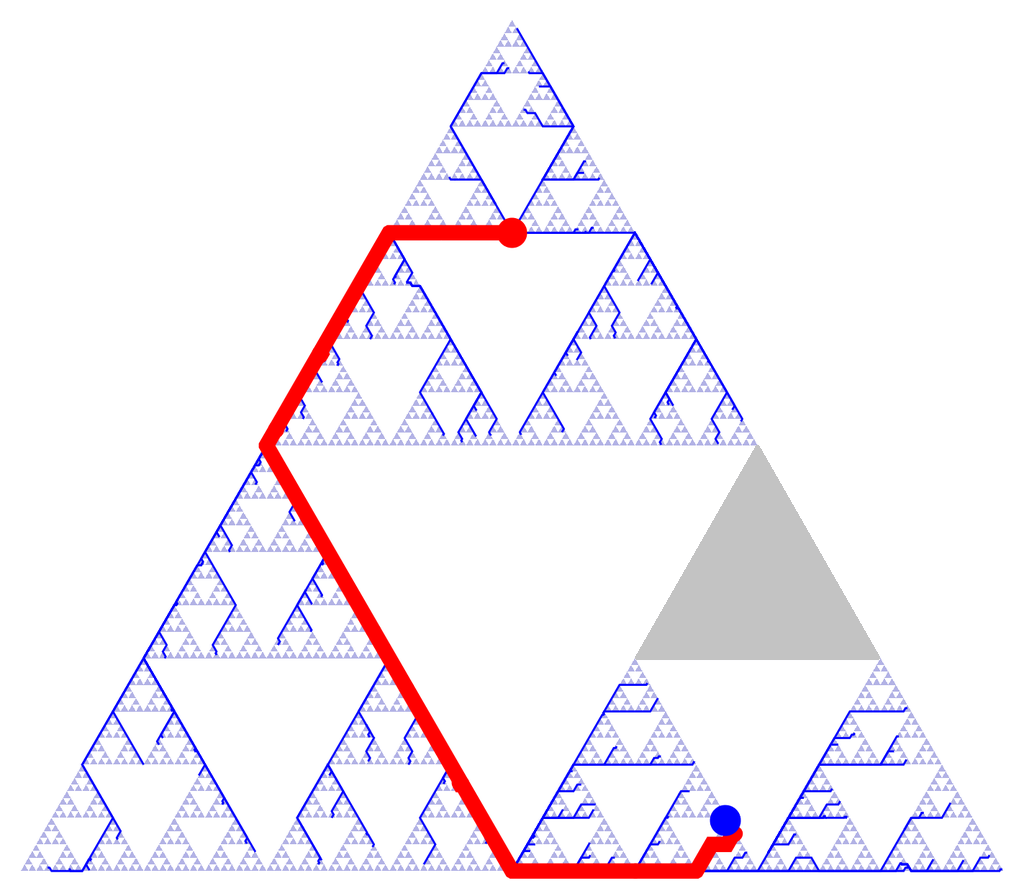}
    \caption{RRT* search tree (blue) and near-optimal trajectories (red) in the Sierpinski gasket $\Sierp$, after 200 iterations. The top figure shows unobstructed trajectory-finding. The bottom figure's grey region indexed by $(2,0)$ was obstructed.}
    \label{fig:Sierpinski}
\end{figure}

We conclude that any search problem on $\Sierp$ implementing the above data satisfies Assumptions \ref{ass:basic} and obtains the convergence guarantees of Theorem \ref{thm:generalized} with $Q=\frac{\log 3}{\log 2}$. We show the results of two RRT* searches in $\Sierp$ in Figure~\ref{fig:Sierpinski}. 

\section{Summary and Conclusions}\label{sec:conclusions}
We provided a new, broader framework for trajectory planning using sampling-based algorithms, including PRM*, RRT, and RRT*, and proved the probabilistic completeness and asymptotic optimality of these algorithms under our broader assumptions. We then showed that multiple systems can be combined into a single larger system using a coupling parameter $p$, in a way that maintains compatibility with the framework and the resulting convergence guarantees.

We demonstrated the framework in three settings. First, we worked with simple linear robots, illustrating  the effect of the coupling parameter $p$ on the motion planning problem results, including some undesirable side-effects of the choices $p=1$ and $p=\infty$. We then applied the algorithms to the task of multi-trajectory planning for multiple car-like robots by combining several Reeds-Shepp vehicle state spaces into a single metric space using an $\ell^p$ distance. We showed that the resulting multi-trajectories depended on the parameter $p$, providing either lowest-total-length multi-trajectories for small values of $p$ or fastest-time multi-trajectories for high values of $p$, with $p=2$ providing convenient intermediate parameter. We finished with an illustration of our framework in the highly non-Euclidean setting of the Sierpinski gasket fractal.

The next step in the development of the method is to extend the framework to allow the minimization of costs that are not given by lengths of trajectories. As was illustrated in Figure \ref{fig:manhattan}, motion planning is very sensitive to the particular choices of data, including the specific choice of $\Extend$ function, and it is unclear at the moment what conditions on the cost functional are sufficient to guarantee convergence.

\bibliographystyle{ieeetr}
\bibliography{valetrefs}

\end{document}